\documentclass[11pt,a4paper]{article}
\usepackage{times,latexsym}
\usepackage{url}
\usepackage[T1]{fontenc}

\usepackage{comment}
\usepackage{graphicx}
\usepackage{amssymb}
\usepackage{amsmath}
\usepackage{booktabs}
\usepackage[table]{xcolor}
\definecolor{qwenpurple}{HTML}{A78BFA}   % your purple
\definecolor{llavaorange}{HTML}{FACC5A}  % your yellow/orange
\colorlet{qwenrow}{qwenpurple!20}
\colorlet{llavarow}{llavaorange!25}
\usepackage{makecell}
\usepackage{arydshln}
\usepackage{multirow}
\usepackage[breakable, skins]{tcolorbox}
\usepackage{float}

\usepackage[acceptedWithA]{tacl2021v1}
\usepackage{xspace,mfirstuc,tabulary}

\newif\iftaclinstructions
\taclinstructionsfalse % AUTHORS: do NOT set this to true
\iftaclinstructions
\renewcommand{\confidential}{}
\renewcommand{\anonsubtext}{(No author info supplied here, for consistency with
TACL-submission anonymization requirements)}
\newcommand{\instr}
\fi

\iftaclpubformat % this "if" is set by the choice of options

\else

\fi

\newcommand\kmnote[1]{\textcolor{red}{\textbf{KM:} #1}}
\newcommand\manote[1]{\textcolor{blue}{\textbf{MA:} #1}}
\newcommand\esnote[1]{\textcolor{orange}{\textbf{ES:} #1}}

\newcommand\aanote[1]{\textcolor{purple}{\textbf{AA:} #1}}
\newcommand\esc[1]{[\esnote{#1}]}

\title{
Does AI See like Art Historians?  \\
Interpreting How Vision Language Models Recognize Artistic Style}

\author{
\textbf{Marvin Limpijankit}$^{1}$,
\textbf{Milad Alshomary}$^{1}$,
\textbf{Yassin Oulad Daoud}$^{2}$ \\
\textbf{Amith Ananthram}$^{1}$,
\textbf{Tim Trombley}$^{2}$,
\textbf{Emily L. Spratt}$^{2}$,
\textbf{Anna Filonenko}$^{2}$,
\textbf{Hannah Pivo}$^{2}$ \\
\textbf{Elias Stengel-Eskin}$^{3}$,
\textbf{Mohit Bansal}$^{4}$,
\textbf{Noam M. Elcott}$^{2}$,
\textbf{Kathleen McKeown}$^{1}$ \\
\\
$^{1}$Columbia University, Department of Computer Science \\
$^{2}$Columbia University, Department of Art History \& Archaeology \\
$^{3}$University of Texas at Austin \ $^{4}$UNC Chapel Hill
}

\date{}

\begin{document}
\maketitle
\begin{abstract}

%KM-TACL - removed unnecessary comments to make it easier to edit

%\esc{when phrased this way, it does point to a natural ``mitigation/solution'' experiment, where you remove concepts that hurt prediction. }

%Also for the final analysis, it will be great to show all 3 directions --> (1) agreement: cases where they find that a model’s use of concepts corresponds to the information they would use in making a judgment on style; (2) disagreement + model is wrong: cases where they find that a model’s use of concepts mismatches with the information they would use in making a judgment on style and is incorrect; but also hopefully some cases of (3) disagreement + model is correct: cases where they find that a model’s use of concepts mismatches with the information they would use in making a judgment on style BUT the model is still using some surprisingly correct/useful/interesting info that humans had not thought of!}

VLMs have become increasingly proficient 
at 
%a range of 
computer vision tasks 
%such as visual question answering and object detection. This 
%including  
%increasingly strong capabilities 
in the art domain.
%from analyzing artwork to generation of art. 
In an interdisciplinary collaboration between computer scientists and art historians, we characterize the mechanisms underlying VLMs’ ability to predict 
artistic style and assess the extent to which they align with
%with the criteria 
art historians. 
%use to reason about artistic style. 
We design a novel approach to latent-space decomposition tailored to the diffuse nature of art style, grounding full image predictions in easy-to-interpret patch level concepts.  This results in a set of  
%to identify 
concepts that drive  
art style prediction and serve as the basis for our
%conduct 
quantitative evaluations, causal analysis and assessment by art historians.
%KM This can be shortened. See my version below. 
%Our findings indicate that  73\% of the extracted concepts are judged by art historians to exhibit a coherent and semantically meaningful visual feature and 90\% of concepts used to predict style of a given artwork were judged  relevant.
Our findings indicate that art historians judge the majority of extracted concepts as coherent and semantically meaningful and also judge the majority of concepts used to predict style of a given artwork as relevant. 
%In cases where 
%When the model used a concept judged irrelevant to successfully predict style, art historians analyzed  possible reasons for its use, noting that the model was able to disentangle  content, form and style  aspects of a concept differently in different contexts. 
In cases of model-expert misalignment, art historians noted the model's use of individual concepts in representing diverse imagery, suggesting it makes connections between different aspects of a concept and different styles.
%(e.g.,  the model might ``understand'' a concept more formally using dark/light
%contrasts). 
% \esc{imo ``exhibit a representation'' is a rare n-gram, i'm used to reading something like ``representations exhibit some quality''. maybe better to say that the extracted concepts exhibit traits, or align with human concepts?}

%\mbc{as i mentioned in my older comments, the 70\% part will sound a buit boring/obvious, can we say a bit more about some surprising/interesting errors historians found (and also some `correct' AI predictions that historians didn't use in their analysis/predictions and were actually surprised by)?}\manote{Yes: This should change to more insightful results once we have the art historians' input}

\end{abstract}

% Submission-specific rules

\section{Introduction}
% \mbc{we need a figure to show the motivating art example + concepts + causality?} (added by Milad)
Humans understand art style by considering both local features, such as texture and color, as well as global properties like composition \cite{barnet2015short}. As vision-language models (VLMs) steadily improve 
%approach human-level performance 
on tasks such as object detection and visual question answering \cite{kim2025visual, zou2023object}, the fundamental question of how models process visual inputs and generate their responses becomes increasingly relevant. This understanding is critical to model interpretability, and the internal mechanisms and decision-making processes of VLMs 
remains an open and challenging 
research 
area \cite{lin2025survey}.

\begin{figure*}[t]
    \centering
    \includegraphics[width=0.95\linewidth]{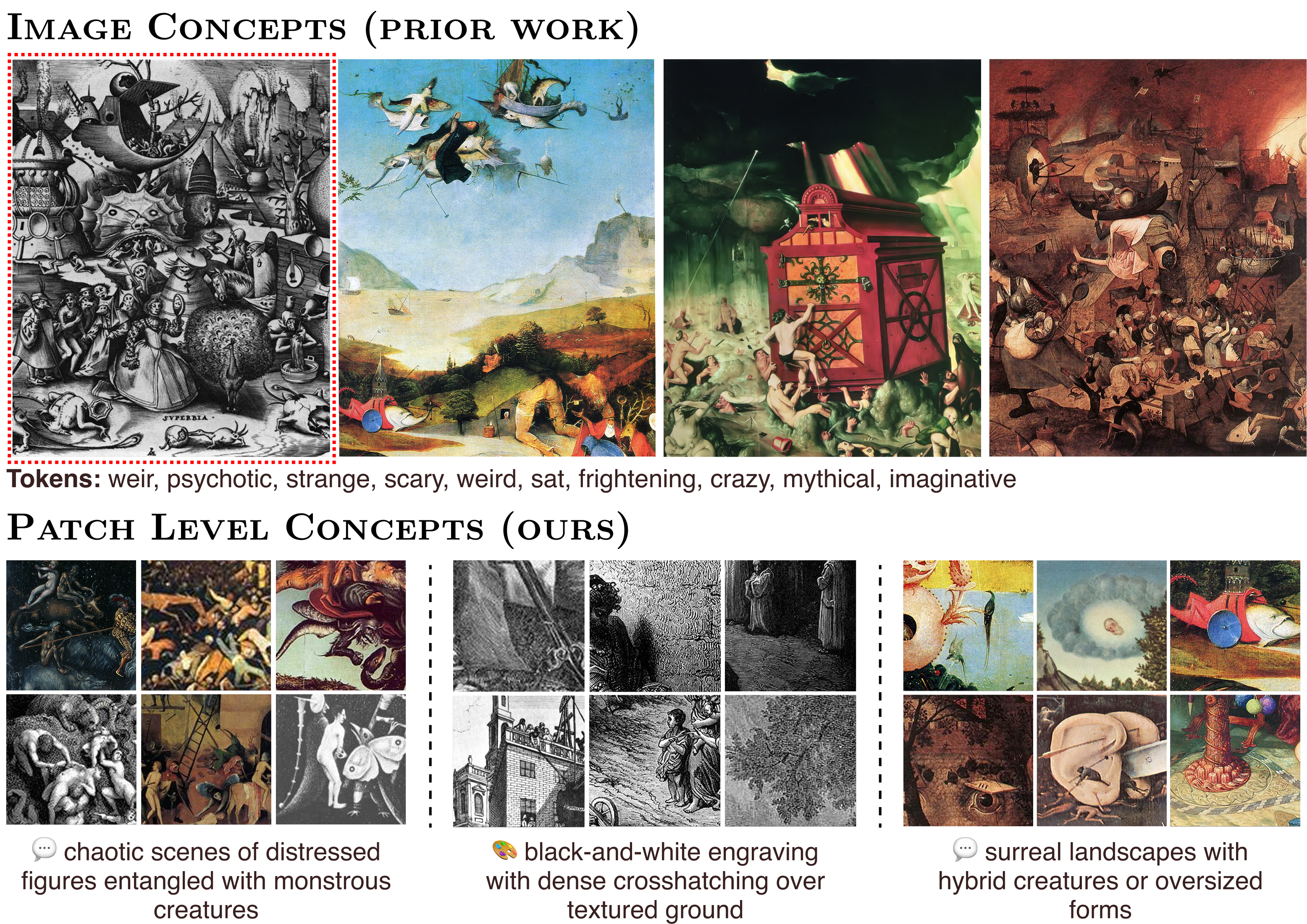}
    \caption{A motivating example. %\kmnote{Could we choose a better example? In this one, the first concept is interior scences, but this is not an interior scene. } 
    \textbf{Top:} The input image (red outline) is classified as Northern Renaissance by the VLM,
    %\mbc{but `Renaissance' not mentioned anywhere in the top part of the figure/caption/tokens? or maybe the caption is outdated...also the bottom middle patch concept says `dense crosshatching over textured background' but is that present in any of the top images? we should try to use the same images for top and bottom parts?}
    however, image concepts offer little explanatory insight---they capture the general motif via visually similar images but do not reveal \textit{why} the model made this prediction. Relevant tokens obtained through logit lens are similarly non-descriptive. \textbf{Bottom:} Our method further resolves image concepts into interpretable, patch-level concepts that distinguish between content and form elements. Here, patch-level concepts are represented by the top 6 image patches (taken from various artworks) that most strongly activate the concept.
    }
    %\aanote{Really minor but "prior" in the top image had me looking for a statistical prior; perhaps "prior work"?}
    \label{fig:our_approach}
\end{figure*}

Identifying art style is particularly challenging for VLMs as these images often contain many rich, fine-grained details \cite{strafforello2025have}.  Visual style is also complex, lacking explicit grounding compared to tasks like object recognition, and models often fail to generalize, relying on patterns from pre-training data rather than faithfully reasoning over the input image \cite{bin2024gallerygpt}. 
%\mbc{add some more motivation/examples from our proposal here of what kinds of location/rural/gender/race aspects we can uncover?} \kmnote{These are not relevant here. I'm adding some text pointing to the figure with examples there. Must change example when Marvin puts new figure in. I see you describe this in detail later so I'm shortening here.  } 
Figure~\ref{fig:our_approach}, for example, shows a complex scene of many figures, where details of its content as well as its form (e.g., black-and-white engraving, crosshatching) contribute to classification as Northern Renaissance.
% people, where details of its content  as well as its form (e.g., smooth brushwork, soft rendering of drapery, sepia tones) contribute to classification as Renaissance.
While prior work has focused on whether VLMs can accurately classify art style, less is known about what visual features drive their prediction and whether these features align with domain knowledge. This raises the following question: \emph{have the models learned to see like human experts, or does their vision operate according to different patterns and logical structures?}
%---perhaps even a fundamentally nonhuman worldview?} 
To investigate this, we focus on the following research questions:
% In this interdisciplinary collaboration involving art historians and computer scientists, we take a model interpretability approach to investigate the following research questions:
\begin{description}
\item[\textbf{RQ1}] What visual concepts do VLMs rely on when predicting %the art 
artistic style?
\item[\textbf{RQ2}] Are the visual concepts causally related to predictions that VLMs make? 
% Do these concepts reflect %visual properties criteria that art historians use when analyzing artistic style?
\item[\textbf{RQ3}] Do visual concepts identified by VLMs align with how art historians see style? 
% What kind of misalignment is there between VLMs and art historians, and how does this affect style prediction?
\end{description}

\begin{figure*}
    \centering
    \includegraphics[width=0.95\textwidth]{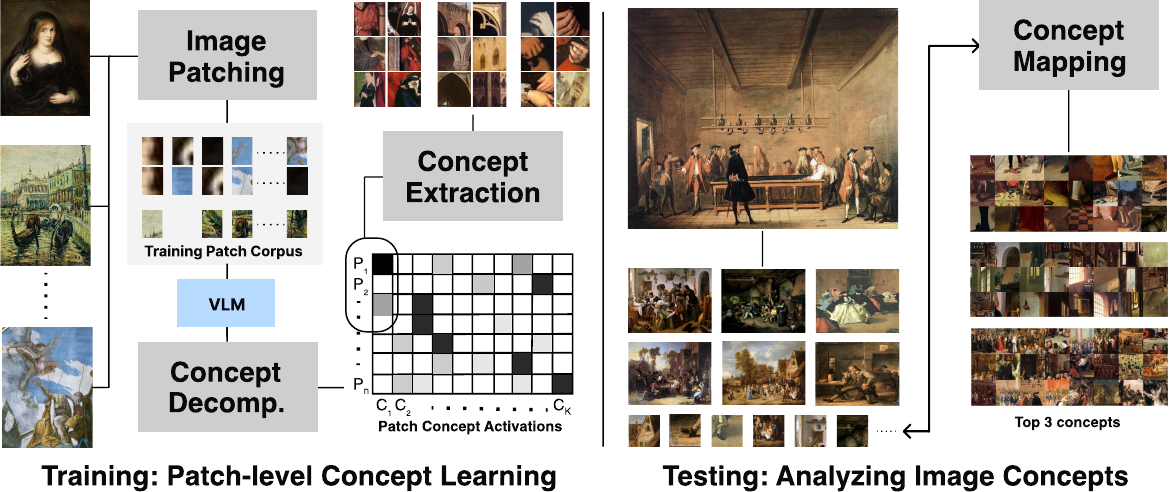}
    \caption{Overview of the concept decomposition pipeline. \textbf{Left:} In the training stage, we (1) split training images into 4x4 patches, (2) pass them to a VLM for style prediction, (3) extract their latent representations and decompose them to obtain patch-level concept activations, and (4) visualize each concept by top activating examples. This process is repeated using full images and a smaller concept size to obtain image-level concepts. \textbf{Right:}  At test time, given an image, we identify its corresponding image-level concept and leverage an image-to-patch concept mapping to retrieve the patch-level concepts most likely to be represented within the image.}
    \label{fig:pipeline}
\end{figure*}

\noindent We address these questions by evaluating
%first j
%open- and closed-source 
VLMs on style classification across %classic 
early modern art, modern art, and architecture datasets.
% designed 
Then, through a novel approach designed to address the particular challenges of interpreting art style, we adapt the full image concept decomposition framework of \citet{parekh2024concept} (previously implemented for object classification). Our approach is unique in its joint decomposition of full images and constituent image patches, grounding full image motifs in more interpretable, patch-level concepts that disentangle the visual interplay of content and form.
%Then, we extend the concept decomposition framework of \citet{parekh2024concept}, which was previously implemented for object classification, to extract visual concepts from the latent representations that VLMs use when classifying style.
% Finding that \texttt{Qwen3} performs strongly, we extend the concept decomposition framework of \citet{parekh2024concept}, which was previously applied to object classification, to the art domain, extracting the task-relevant visual concepts that VLMs use when predicting style. 
%We augment the approach with image patching to recover interpretable concepts that disentangle the visual interplay of content and form.
%We develop a novel approach for connecting image patches with the full image and 
%We develop a novel approach involving running concept decomposition on both the patch and full image level and grounding the full image motifs in more atomic, interpretable, patch-level concepts that disentangle the visual interplay of content and form.
%and introduce a more descriptive concept labeling method (Figure~\ref{fig:pipeline}).
%new concept labeling approach that produces richer, more descriptive concept tags (Figure~\ref{fig:pipeline})
% that characterize different artistic styles (Figure~\ref{fig:pipeline}).
% A key feature of our approach is the integration of localized image patches in our interpretability framework in order to disentangle complex visual interplay of content and form that characterize different artistic styles. 
% This process is outlined in Figure~\ref{fig:pipeline}.
We train a linear probe to identify patch concepts that the model strongly associates with certain styles, and subsequently confirm through intervention tests that masking these patches causally impacts full image style prediction more than random patches.
% We perform intervention experiments to validate the causal impact of concepts on style prediction and supplement this with a linear probe analysis to identify style-specific concepts. 
% complement this with a correlational analysis to identify style-specific concepts.
%confirming that these concepts causally affect the style prediction, and supplement this with a correlational analysis based on linear probing to identify the most style-relevant concepts.

%Following this, 
To understand whether these patch level concepts align with human analysis of style, 
our research involves collaboration with six art historians, comprising graduate students and faculty, who provide expert
%ise
analysis of the computational results. % to conduct two qualitative studies. 
In one qualitative study, experts perform an intrinsic evaluation of concept quality; they examine and label each concept, revealing that the concepts are meaningful and reflect a range of thematic dimensions from content-based features (e.g., objects and scenes) to form-based features (e.g., color palette, texture, and lighting).
In a second extrinsic study, experts analyze %test 
images together with the model's prediction of art style and most associated patch concepts. We find that the concepts that were most associated with artwork, according to the model,  were  judged relevant in the majority of cases to both the image and predicted style. % \mbc{but don't we also need a discriminative/causal study where we show historians patches and correct vs. incorrect images and check for some predictiveness of our learned patches? i.e. the human study version of ` `causal and correlational analysis demonstrating that concepts causally influence style classification performance'?}
% \esc{I agree w/ this, i wonder also if it would be useful to have the art historians say what they use to make their classification that's *not* captured by the patches. The patches that the model uses are relevant, but are they sufficient? it seems like there would be interesting things to talk about there in terms of motifs/patterns that are outside of the patch level, but that humans can pick out easily}
Notably, cases where expert judgment diverged from the model proved particularly revealing, suggesting the model captures visual cues that, while predictively useful, fall outside conventional art-historical categorization. 

Our contributions, which provide answers to the corresponding research questions, are:
\begin{enumerate}
    \item An approach for decomposing VLM art style predictions into interpretable patch-level visual concepts.
    %identifying interpretable visual concepts that VLMs use for art style classification by grounding image predictions in interpretable patch-level visual concepts.
    %using image patches and a method for connecting them to full image prediction.
    \item A correlational and causal analysis that demonstrates how VLMs' art style predictions are grounded in patch-level visual concepts.
    %demonstrating that patch-level visual concepts causally influence style classification performance
    \item A framework for measuring alignment between VLMs and art historians for art style classification revealing that there is considerable agreement on the relevance of patch-level visual concepts. The analysis also provides clues about ways in which models and art historians rely on different visual features for art style classification. 
    % An extension of VLM concept decomposition to art style classification that operates at the patch level and a method for identifying relevant concepts to full image predictions
     %An interpretable framework for VLMs 
%    what is unique about the method? Go beyond developing the pipeline to enable analysis of each stage in the pipeline. 
  %  \item An interdisciplinary comparison between the model's stylistic analysis and the canonical analysis of art historians revealing that most identified concepts are relevant to prediction of style as well as explanations for why concepts judged irrelevant might nonetheless result in successful style prediction. 
\end{enumerate}

\section{Related Works}

% art analysis (whether models can do style classification -> how models do style classification)

% Karayev, 2013 - Recognizing Image Style \cite{karayev2013recognizing} (CNNs for style) [X]
% Xu, 2014 (architecture style classification) \cite{xu2014architectural}
% Lecoutre, 2017 - Recognizing Art Style in Painting \cite{lecoutre2017recognizing} (CNN, transfer from ImageNet to art style) [X]
% Garcia, 2018 - How to read paintings (semantic understanding retrieval, can potentially skip to focus on style here) \cite{garcia2018read}
% Garcia, 2019 - context-aware embeddings for automatic art analysis (enhances embeddings with artist information, can potentially skip too) \cite{garcia2019context} 
% Garcia, 2020 - VQA for art (focus on visual details and knowledge but assesses style too) \cite{garcia2020dataset} [X]
% Menis-Mastromichalakis, 2020 \cite{menis2020deep} (ensemble approach, representations from multiple models for style detection) [X]
% Castellano & Vession, 2021 \cite{castellano2021deep} (survey on DL for art) [X]
% Bleidt, 2024, ArtQuest \cite{bleidt2024artquest} (fixes language bias in ArtVQA) [X]
% Bin, 2024, GalleryGPT \cite{bin2024gallerygpt} (formal analysis generation) [X]
% Strafforello, 2025 (have large vision-language models mastered art history?) \cite{strafforello2025have}

Automatic recognition of artistic style has been studied for over a decade \cite{karayev2013recognizing,castellano2021deep}.  While early approaches relied on hand-crafted visual features such as color histograms \cite{li2009aesthetic}, since then, learned local and global features of convolutional neural networks (CNNs) have yielded substantial improvements in accuracy \cite{lecoutre2017recognizing,menis2020deep}.  The strength of these representations have expanded the frontier of tractable domains beyond paintings to architecture \cite{xu2014architectural} and recent advances in vision-language models have made open generation tasks such as question answering and formal analysis possible \cite{garcia2020dataset,bleidt2024artquest,bin2024gallerygpt}.  A recent evaluation of state-of-the-art VLMs on recognition of artistic style revealed impressive capabilities while also highlighting systematic disagreements with ground-truth labels that may reflect contested human categorizations \cite{strafforello2025have}.  Our work sheds light on this question by shifting the focus from \textit{whether} VLMs can classify style to \textit{how} they do so, comparing the mechanisms underlying VLM style predictions directly against the judgments of art historians.  

We do so by leveraging methods for model interpretability.  These techniques, which include linear probing \cite{alain2016understanding}, activation patching \cite{wang2022interpretability}, dictionary learning \cite{lee1999learning,olah2020zoom,fel2023craft} and sparse autoencoders \cite{bricken2023monosemanticity,cunningham2023sparse}, aim to explain model decisions by decomposing dense network activations into human-understandable concepts.  \citet{parekh2024concept} extend these approaches to vision-language models by showing that Semi-Nonnegative Matrix Factorization (Semi-NMF) can yield concepts that are grounded in both visual and textual modalities.  We adapt their framework to visual art via a novel %\kmnote{Can we insert the word "novel" here? }
patch-level decomposition that localizes visual features, informed by the spatially distributed nature of stylistic signals.  Crucially, our work pairs this computational approach with interdisciplinary inquiry, contributing to an emerging literature that measures model alignment with categories that human experts actually use \cite{orgadinterpretability}. In doing so, we act on prior calls for closer dialogue between computational image analysis and humanistic frameworks, which have more to offer each other than either discipline has fully realized \cite{spratt2014computational,spratt2017dream}.

% \aanote{There are many more papers on interpretability (and VLM interpretability) that I could include but as our contribution isn't a new interpretability method I thought it better not to -- in particular, I was worried about priming reviewers to read the paper as a new interpretability technique. Happy to add it if others disagree.}

\begin{comment}

\section{Data}

%\kmnote{For future reference, all numbers less than 10 should be spelled out in words. I have chnaged}
We curate three datasets each consisting of images from five style categories. Two datasets are sourced from WikiArt,\footnote{\url{https://www.wikiart.org}} which focuses on artworks, whereas the third is from the Architecture dataset, which contains images of architectural styles \cite{xu2014architectural}. We used 2,500 images for each artwork dataset and 1,500 images from the architecture dataset sampled evenly across styles. As detailed in \S\ref{sec:discovering_concepts}, these images are then split into 16 patches each, resulting in 40,000 and 24,000 samples respectively. Table \ref{tab:dataset_summary} summarizes the styles in each dataset. We group styles that share visual characteristics and belong to the same broad category, creating a challenging fine-grained classification task.

\end{comment}

\section{Methods}

\subsection{Data}

%\kmnote{For future reference, all numbers less than 10 should be spelled out in words. I have chnaged}
We curate three datasets each consisting of images from five style categories. Two datasets are sourced from WikiArt,\footnote{\url{https://www.wikiart.org}} which focuses on artworks, whereas the third is from the Architecture dataset, which contains images of architectural styles \cite{xu2014architectural}. We used 2,500 images for each art dataset and 1,500 images from the architecture dataset sampled evenly across styles. As detailed in \S\ref{sec:discovering_concepts}, these images are then split into 16 patches each, resulting in 40,000 and 24,000 samples respectively. Table \ref{tab:dataset_summary} summarizes the styles in each dataset. We group styles that share visual characteristics and belong to the same broad category, creating a challenging fine-grained classification task.

\begin{table}[t]
\centering
\small
\setlength{\tabcolsep}{4pt}
\renewcommand{\arraystretch}{1.15}
\begin{tabular}{p{0.225\columnwidth} p{0.7\columnwidth}}
\hline
\textbf{Dataset} & \textbf{Styles} \\
\hline
WikiArt (Early Modern)
& Baroque, Northern Renaissance, Realism, Rococo, Romanticism \\
\hline
WikiArt (Modern)
& Abstract Expressionism, Color Field, Cubism, Fauvism, Minimalism \\
\hline
Architecture
& Art Nouveau, Baroque, Byzantine, Gothic, Romanesque \\
\hline
\end{tabular}
\caption{Datasets and associated style categories used in our experiments.}
\label{tab:dataset_summary}
\end{table}

\subsection{Discovering Concepts}
\label{sec:discovering_concepts}

The goal of concept decomposition is to learn a set of interpretable concepts that capture recurring patterns in model representations. This is often achieved via dictionary learning based methods \cite{bussmann2024batchtopk, rajamanoharan2024jumping}. In particular, Semi-Nonnegative Matrix Factorization (Semi-NMF) has been shown to effectively find concepts that VLMs associate with identifying specific objects \citep{parekh2024concept}. We extend this approach to art style prediction. %\esc{these citations should be propagated into the intro}

Formally, for a dataset of images $\mathcal{X} = \{x_1, \dots, x_n\}$ with corresponding ground truth style labels $y = \{y_1, \dots, y_n\}$, we prompt a VLM to classify each image $x_{i}$ as one of five style categories. Following \citet{parekh2024concept}, we extract the residual-stream representation from a specified layer $L$ upon generating the first token of a target word, specifically the model's style prediction (e.g., `Baroque'). The result is a matrix $\mathbf{Z} \in \mathbb{R}^{d \times n}$, where the $i$-th column contains the $d$-dimensional latent representation used by the VLM to predict the style of image $x_i$. This matrix is then decomposed ($\mathbf{Z} \approx \mathbf{U} \mathbf{V}$) under the optimization,
\begin{equation}
\begin{aligned}
\mathbf{U}^*, \mathbf{V}^*
&= \arg\min_{\mathbf{U},\,\mathbf{V}}
\|\mathbf{Z} - \mathbf{U}\mathbf{V}\|_F^2 + \lambda \|\mathbf{V}\|_1
\\
&\makebox[0pt][r]{\text{s.t. }} \mathbf{V} \ge 0,\;
\|\mathbf{u}_k\|_2 \le 1,\;
\forall k \in \{1,\dots,K\}
\\[1ex]
\end{aligned}
\end{equation}

\noindent where $K$ is a pre-specified number of concepts, $\mathbf{U} \in \mathbb{R}^{d \times K}$ is the learned dictionary of concepts (normalized $d$-dimensional vectors), and $\mathbf{V} \in \mathbb{R}^{K \times n}$ contains the concept activations for each image. Activations are strictly nonnegative and the $\lambda$ parameter encourages a few concepts to be active per image (sparsity). 

\textbf{Image Patching.} 
%KM suggested rewording.
Identifying concepts across multiple images of artwork is more difficult than in other genres-such as the photographs used for object recognition-given the many detailed elements and relations that are depicted. 
%\esc{list example genres here, since reader might think genre here refers to genre of art}
%\kmnote{I have done this}
%It is difficult to identify concepts across multiple images of artworks since each contains many complex elements. 
Thus, as in \citet{kondapaneni2025representational}, who examine shared object concepts across models, we first split each image into a $4 \text{x} 4$ grid of patches and then apply concept decomposition at the patch-level (i.e., each input is an image patch). This localizes image features, making it easier to generate interpretable concepts. 
%\esc{this will probably raise questions from reviewers like:
%- how sensitive is the method to the patch size
%- what if you used a more informed approach like bounding boxes?
%- how does this interact with the image tokenizer? } \kmnote{Leaving this in for now as we won't address it by the time of arxiv but we can consider how to address for TACL. }

\begin{comment}

\begin{table}[t]
\centering
\small
\setlength{\tabcolsep}{4pt}
\addtolength{\tabcolsep}{3pt} % <-- horizontal stretch
\renewcommand{\arraystretch}{1.15}
\begin{tabular}{l r r r}
\hline
\textbf{\begin{tabular}[c]{@{}l@{}}Model\\~\end{tabular}} &
\textbf{\begin{tabular}[c]{@{}r@{}}Layer\\~\end{tabular}} &
\textbf{\begin{tabular}[c]{@{}r@{}}Threshold\\Percentile\end{tabular}} &
\textbf{\begin{tabular}[c]{@{}r@{}}Avg.\ Active\\Concepts\end{tabular}} \\
\hline
Llava-1.5 & 20 & 0.60 & 15.64 \\
Llava-1.5 & 20 & 0.80 & 7.82 \\
Llava-1.5 & 20 & 0.90 & 3.91 \\
Llava-1.5 & 30 & 0.60 & 14.67 \\
Llava-1.5 & 30 & 0.80 & 7.33 \\
Llava-1.5 & 30 & 0.90 & 3.67 \\
Qwen3 & 20 & 0.60 & 15.60 \\
Qwen3 & 20 & 0.80 & 7.80 \\
Qwen3 & 20 & 0.90 & 3.90 \\
Qwen3 & 30 & 0.60 & 11.46 \\
Qwen3 & 30 & 0.80 & 5.73 \\
Qwen3 & 30 & 0.90 & 2.86 \\
\hline
\end{tabular}
\caption{Sparsity statistics across models and layers.}
\label{tab:sparsity_table}
\end{table}

\end{comment}

\textbf{Hyperparameters.} We experiment with multiple layers $L$ at which the latent representations are extracted and the number of concepts $K$. 
%KM attempting to shorten by one line when I found this. We already say this earlier so I find it repetitive.
%The goal is that concepts should help explain model outputs. 
We evaluate configurations using linear probe accuracy (see \S\ref{sec:concept_style_associations}), which measures how well concept activations predict the model's output style, and consider additional metrics in Appendix \ref{sec:appendix_parameters}.
%Additionally, we investigate concept homogeneity, the entropy of style predictions among images that activate a concept, in [APPENDIX].

\textbf{Activation Threshold.} Semi-NMF encourages sparsity but results in many near-zero activations that can create noise. Therefore, to determine an activation threshold, we experiment with different percentiles and examine the number of activated concepts per patch.
%(Table \ref{tab:sparsity_table}).
Following this, we select a threshold at the 90th percentile of non-zero activation values as this yields two-three activated concepts on average (Appendix \ref{sec:appendix_parameters}), reflecting our expectation that small patches should correspond to only a few concepts.

\textbf{Concept Representation.}
To interpret the discovered concepts, we represent them using prototyping \cite{alvarez2018towards, kim2018interpretability}, displaying the top $k=24$ image patches\footnote{Throughout this paper, we display a smaller number of image patches per concept due to space constraints.} with maximal concept activation as shown in the human study interface in Appendix \ref{sec:appendix_annotation_guidelines}.
%\footnote{In our experiments, we show top 16 image patches}. 
Moreover, we experiment with four methods of generating text labels that describe the concepts
%' top-activating images 
(see Appendix~\ref{sec:appendix_concept_labeling_methods}).
%\manote{The appendix doesn't explain what the labeling methods are! I couldn't find anywhere where we explained them!}.
Example generated labels are shown in Figure \ref{fig:our_approach}. These labels can be useful in studies where it is not possible to create manual labels by human experts.

\subsection{Concept-Style Associations}
\label{sec:concept_style_associations}

We assess VLMs' concept-style associations via probing, where a linear classifier is trained to predict the model's response (i.e., the predicted style) based only on concept activations---a single vector indicating which concepts the image activated. Finding that this classifier identifies the model's style prediction reliably (see \S\ref{sec:concept_analysis}), we inspect the learned weights of the classifier and extract the most positively associated concepts for each style.

\subsection{Causal Analysis via Intervention}

We investigate the causal impact of patch-level concepts on full image style prediction by testing whether ablating them degrades performance. 
Namely, for each art style $y$, we collect samples where the model correctly classified the image $\mathcal{D}_y$.
Let $\mathbf{x} \in \mathcal{D}_y$ be one such example, we calculate the log probability of the model predicting the target style $\log p_{\theta}(y \mid \mathbf{x})$ over the first generated token as a baseline. Then, we consider three masking interventions: (1) we mask the $N=4$ patches with the highest classifier weights (from \S\ref{sec:concept_style_associations}) for the target style, (2) we randomly mask $N$ patches, and (3) we randomly mask $N$ patches excluding those in (1). Specifically, masked patches are replaced with a Gaussian blur of radius $r=20$. As (2) and (3) are randomized, we repeat each 10 times and average the result. We report the average impact of each intervention on a target style as
\begin{equation}
\Delta_{\text{int}, y} = \frac{1}{|\mathcal{D}_y|}
\sum_{\mathbf{x} \in \mathcal{D}_y}
\bigl[ \log p_{\theta}(y \mid \tilde{\mathbf{x}}_{\text{int}})
- \log p_{\theta}(y \mid \mathbf{x}) \bigr]
\end{equation}
where $\text{int} \in \{1, 2, 3\}$ is an intervention and $\tilde{\mathbf{x}}_{\text{int}}$ is the corresponding intervened image.
Assuming these patch-concepts are causally important to the model's prediction of style $y$, we expect $\Delta_{1, y} < \Delta_{2, y} < \Delta_{3, y}$.

\subsection{From Patches to Images}

Patch concepts correspond to more local and interpretable features, but connecting them back to 
%understand model predictions on 
full images is challenging due to the distribution mismatch between their latent representations (visually simple patches vs. full-resolution artworks).
%KM I think you can remove this. Obvious. 
%which we address in this section. 
\begin{comment}
\esc{isn't this the method you are introducing? if it is, then it feels odd to say ``underexplored'' here, since it implies something about prior work} \kmnote{I do feel the paper tends to read "based on X" which raises the question of how much is novel. So I agree with Elias here that highlighting it as your work would be better. }

\esc{if this is a finding of yours, i would say something like ``However, we find that this leads to...'' to make it clear}
\end{comment}
% One approach is to similarly extract the VLM's latent representation of the full image and directly apply the concept decomposition learned at the patch level. However, we find that this leads to non-sparse activations and certain concepts activating on nearly all images, producing incoherent results. %[APPENDIX]
% We attribute this to the domain shift from patches, which are lower quality and visually simple, to full-resolution images of artworks. 
% \kmnote{I feel that this section is long and it's not clear to me that anything other than the first sentence is needed in the above paragraph. You are spending space on a method you did not use and which noone else uses either (since I see no cite). Could you go from first sentence into next paragraph? If you feel something is needed then can you reduce the above? }
Observing that image-level concepts capture general motifs composed of multiple, smaller features (Figure \ref{fig:our_approach}), we propose to ground them by identifying their most distinctively associated patch concepts---analogous to finding characteristic words (patch concepts) for a topic (image concept) from a corpus of documents (images) in topic modeling \cite{newman2010automatic}. 
%\esc{IMO this is too broad/awkward because it makes it sound like this paper is not an NLP paper, which then makes it ill-fit for TACL. i would say something more specific like ``topic modeling'' and cite relevant prior work.}
First, we run concept decomposition on both the image and patch level. Then, for each full image concept $c_{f}$, we take all images where $c_f$ is active and represent each as a binary bag-of-patch-concepts---a vector indicating which patch concepts $c_p$ are active in any of its patches. We then count co-occurrences of each $c_p$ across these images and rank by PMI score
\begin{equation}
    \text{PMI}(c_f, c_p) = \log \frac{P(c_p \mid c_f)}{P(c_p)}
    \label{eq:pmi_scoring}
\end{equation}
where $P(c_p \mid c_f)$ is the fraction of $c_f$-activating images containing $c_p$, and $P(c_p)$ is the marginal rate of $c_p$ across all images.
At test time, we map each image to its most strongly activated full image concept and retrieve the corresponding highest-PMI patch concepts. Examples are provided in Appendix \ref{sec:appendix_additional_concept_examples}. We evaluate the effectiveness of this approach in a study presented in \S\ref{sec:art_interpretability}.

%Therefore, we propose the following method to identify the relevant patch concepts from a full image. Given the dataset of images, we run concept decomposition at both the full image and patch level, yielding $K_{full}$ and $K_{patch}$ concepts respectively ($K_{full} \ll K_{patch}$). Then, the patch-level concept activations are binarized using a percentile-based threshold $\tau_{patch}$ computed across all non-zero activations.
\begin{comment}
\esc{I would be prepared to explain how this threshold was chosen, and to show how sensitive the method is to the threshold. But this is lower priority, can happen in the rebuttal.} \kmnote{or in appendix. }
\end{comment}
% This results in a $K_{patch}$-dimensional $\{0, 1\}$ vector for each patch, where the $i$-th entry indicates the activation of concept $i$. We then aggregate patch vectors into a single representation for the full-image by taking their element-wise OR, yielding a $K_{patch}$-dimensional binary vector indicate which concepts were active in any of the image's patches. We apply a similar threshold $\tau_{full}$ for full image concepts, and, using the result of the previous step, compute $P(c^{patch}_i \mid c^{full}_j)$--the probability that a patch concept $i$ is present in the image given it activated full image concept $j$--empirically from co-occurence counts. We experiment with thresholds and set $\tau_{patch}$ to the 95th percentile and $\tau_{full}$ to the 80th percentile of non-zero activations.
%report findings in [APPENDIX]. 
\section{Experiments and Results} %\kmnote{check }
\begin{figure*}[t]
    \centerline{\includegraphics[width=0.95\textwidth]{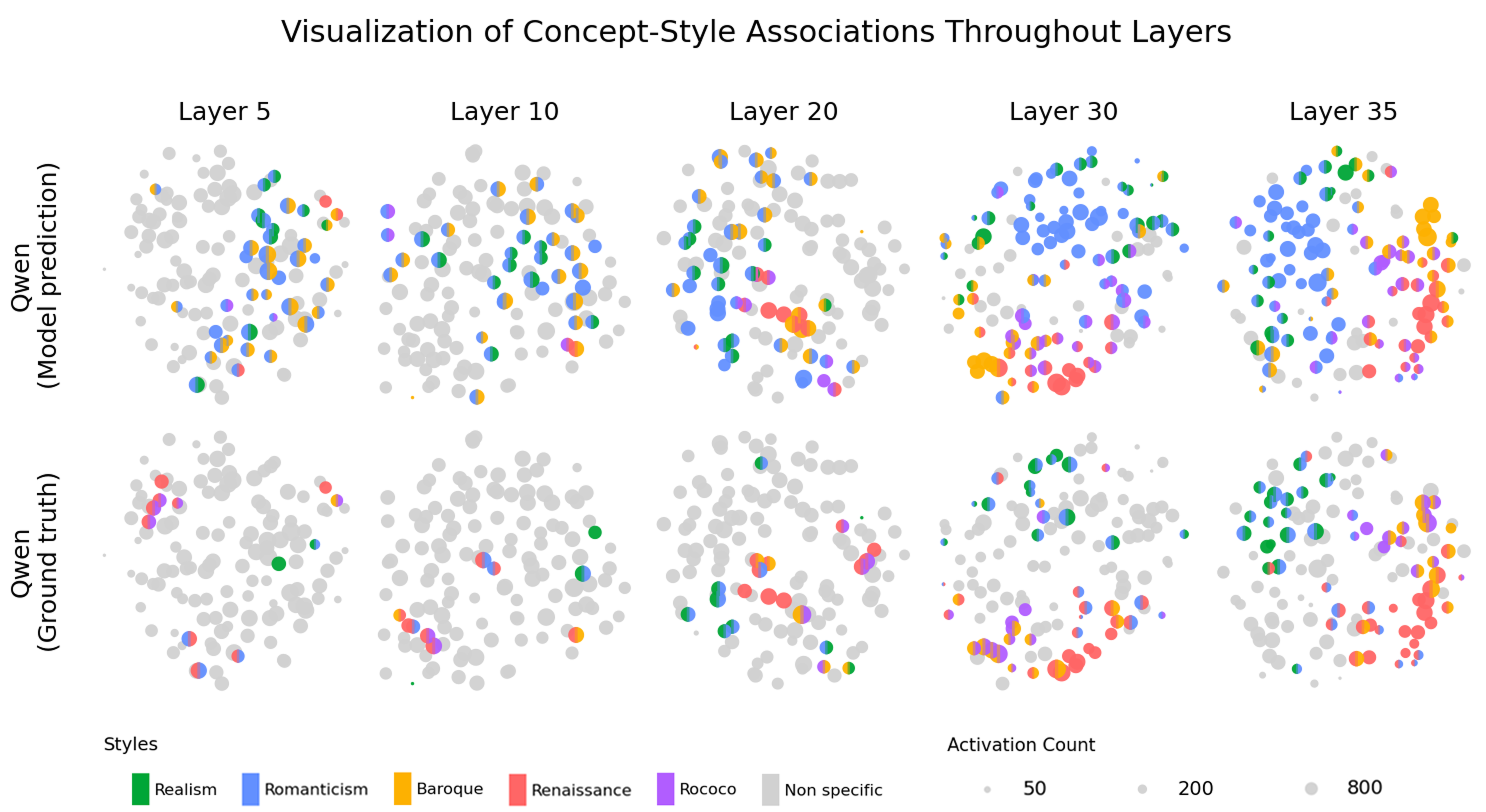}}
    \caption{The space of concepts visualized in 2D via t-SNE. Concepts more similar to each other in high-dimension are closer. The size of each concept indicates how frequently the concept is activated across the dataset. Colors indicate style-specific concepts--instances where $>70\%$ of the activated images correspond to 1 (or if not, 2) styles. These styles either correspond to the model's prediction (top) or the ground truth style (bottom).}
    \label{fig:concept_visualization}
\end{figure*}
\subsection{Model Benchmarking}

\begin{comment}

\begin{figure*}[t]
    \centerline{\includegraphics[width=1\textwidth]{figures/linear_probe.png}}
    \caption{Your caption here}
    \label{fig:linear_probe}
\end{figure*}

\end{comment}

Performance on art style classification across several popular VLMs is included in Appendix \ref{sec:appendix_2_model_benchmarking}. %\ref{fig:art_style_classification}. 
% We introduce a control set of WikiArt images consisting of 5 significantly different styles. 
\begin{comment}
\aanote{In the dataset section, we need to introduce and motivation the selection of our style sets more explicitly}
\end{comment}
\texttt{GPT5} and \texttt{Qwen3} achieve comparable results, with the best accuracies, while \texttt{Molmo2} and \texttt{LLaVA-1.5} exhibit a notable drop-off.
%\kmnote{Would be good to add here that results are comparable }
Models tend to perform better on the architecture dataset relative to WikiArt datasets. 
%[APPENDIX] 
%\aanote{I know we're tight on time but it would be nice to have an expert score for these tasks -- if we delay the submission perhaps we can?  I just think Wikiart in particular has contestable style labels in it.  Alternatively if prior work has multi-annotation human judgment of Wikiart, reporting that would be nice.} 
%\kmnote{We should have this in results of our current study. I believe we asked the art historians to predict styles. I think Noam is also interested in this. Milad - can you include? }
We also introduce a control set with 5 highly distinct WikiArt styles, on which models perform strongly, suggesting that difficulty on WikiArt stems from the use of closely related styles. %\aanote{This is an interesting dimension of this work -- closely related styles -- that we should consider highlighting in the intro.  Perhaps we can tie it into the interdisciplinary dimension?} 
%\esc{agree w/ this -- is there any way to quanitfy how related two styles are, maybe there's existing art history work on this? or could one use temporal information to approximate?}
%rather than just the difference between the architecture and artwork. 
We observe some disagreement among domain experts on the WikiArt style labels (see \S\ref{sec:art_interpretability}), which may partially explain these results. %\cite{strafforello2025have}. 
%\aanote{Oh, do they have ceiling performance numbers we can report here?}
We focus our analysis on \texttt{Qwen3} and \texttt{LLaVA-1.5}, two open-source models with marked differences in accuracy. 
%Appendix~\ref{sec:appendix_model_benchmarking} provides a per-style breakdown of the results.
\begin{comment}
\begin{figure}[h]
    \centering
    \includegraphics[width=\columnwidth]{./figures/art_classification_new.png}
    \caption{Overview of VLM performance on zero-shot art style classification (full image).}
    \label{fig:art_style_classification}
\end{figure}
\end{comment}
\subsection{Concept Analysis}
\label{sec:concept_analysis}
\noindent\textbf{How do models associate concepts with art styles?} Figure \ref{fig:concept_visualization} visualizes the space of concepts (high-dimensional vectors) at different layers in 2D using t-SNE for \texttt{Qwen3}. 
% \esc{mention tsne}
The concepts specific to a certain style, either the model's predicted style or the ground truth, are colored. We observe that as representations progress throughout layers, the model identifies concepts that it associates strongly with %a outputting
a specific style. These, however, do not always reflect their ground truth annotations. For example, concepts in the bottom-right of the layer 35 plot (Renaissance, Rococo, Baroque) align with ground truth labels whereas in the top-left, the model mistakenly associates some concepts with Romanticism when these examples are actually annotated as Realism. However, the art historians in this study also note that the dataset contains overlapping images between Romanticism and Realism, and that these styles are not clearly delineated. %(see \S\ref{sec:art_interpretability}). \kmnote{Check this. I think current version does not include this.}
%\kmnote{Either refer forward to art history discussion or refer backward to this from art history results discussion}
The concepts are broadly similar in activation frequency, indicating a balanced decomposition. 

\begin{figure}[h]
    \centering
    \includegraphics[width=0.98\columnwidth]{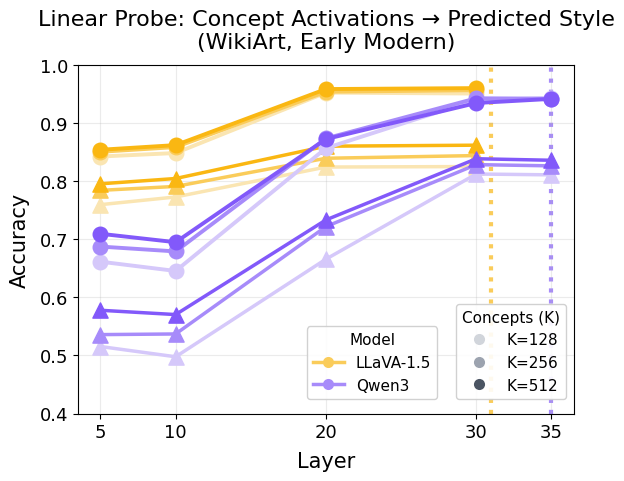}
    \caption{Accuracy of a linear probe trained to predict model output style from its concept activations ($\bullet$ raw activations; $\blacktriangle$ binarized activations).
    %\esc{what about normalizing the x axis here to be a percentage of total layers in the model, so that both models start/end in the same place} 
    }
    \label{fig:linear_probe}
\end{figure}

\textbf{Can concepts predict the model's style classification?} We train concept decomposition models under different configurations and construct a meta-model by fitting a linear classifier to predict the model's style prediction for an image patch based on its concept activations. 
%At test time, we pass unseen image patches to the VLM, extract their latent representations, decompose them into concept activation vectors, and measure the accuracy of the trained probe. 
Figure \ref{fig:linear_probe} shows the linear probe test accuracies on the Early Modern subset of WikiArt.
%Results on other settings are deferred to the appendix due to space constraints.
\begin{comment}
\aanote{I think this style of figure is very cool but wonder, rather than (or in addition to) showing each model separately, perhaps it would be cool to have each row correspond to a single style from Qwen.  Then, we could see how each style's reliance on concepts evolves over time -- ideally, in early layers, styles rely on similar concepts that are then composed differently in later layers.}
\end{comment}
Concept activations from later layers are better able to predict the model's output, reaching up to 0.95 accuracy at the final layer. This is consistent with previous work showing that representations become increasingly task-relevant towards the final layer \cite{wang2022interpretability}.
%\kmnote{"notions" is odd. Can you say "with previous work demonstrating that representations.."?}
Notably, for \texttt{LLaVA-1.5}, even concept activations in early layers are sufficient. This is likely due to this model's tendency to predict one-two styles nearly all the time, contributing to the poor classification performance observed. 
%[APPENDIX]. 
We additionally experiment with using the binary representation of concept activations post thresholding, and find that just these top few activated concepts remains sufficient for predicting model output (0.85). 

\begin{figure}[h]
    \centering
    \includegraphics[width=\columnwidth]{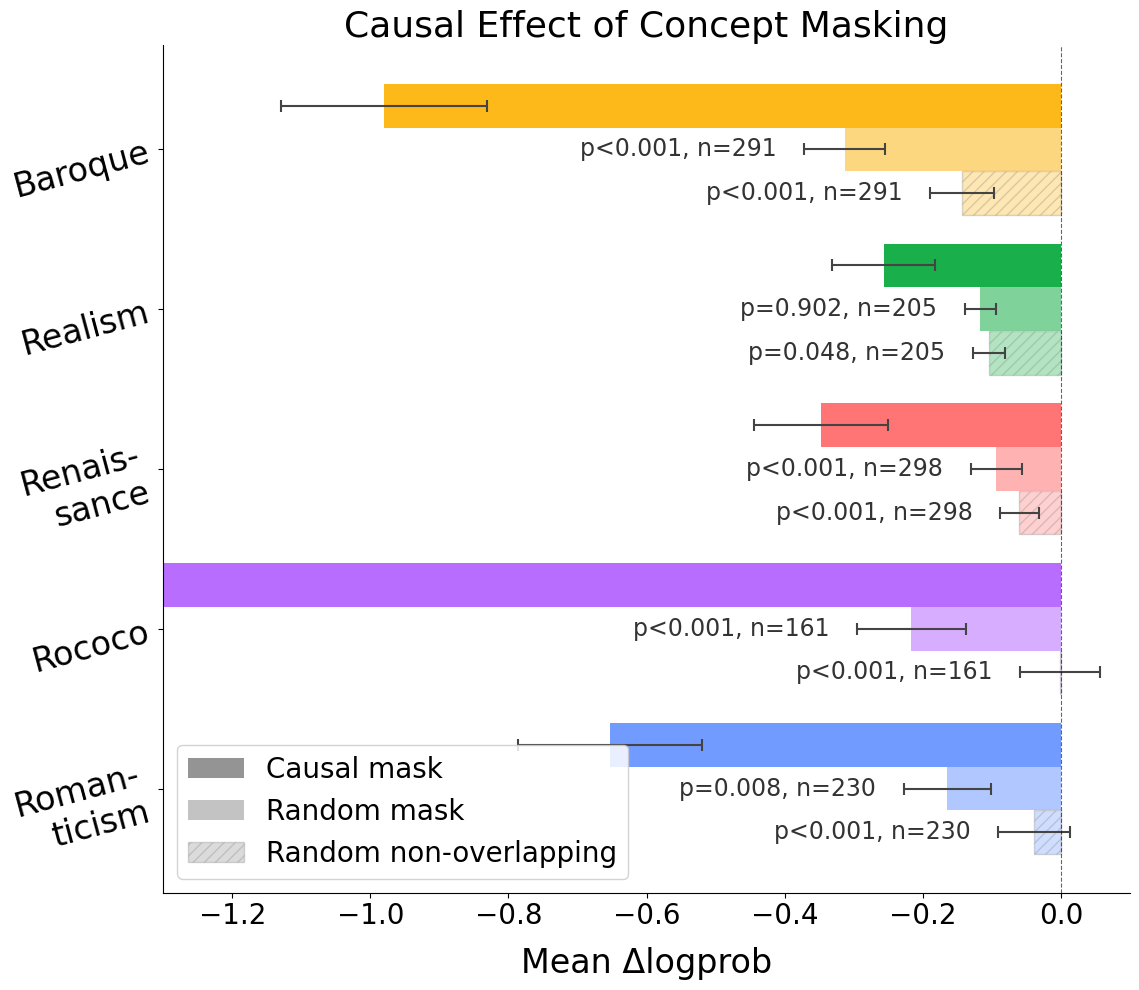}
    \caption{Shift in log-probability of correct style prediction under three masking settings: concept, random, and random non-overlapping.}
    \label{fig:causal_effect_full}
\end{figure}
\subsection{Concept-Style Associations}
\textbf{Which concepts drive the model's prediction of each style?} The weights of the linear probe are used to identify the concepts most correlated with the model prediction of each style. We additionally validate that at the patch-level, these weights align with the causal effect of directly manipulating the model's latent representation in the concept direction (Appendix \ref{sec:appendix_causal_analysis}).  In Figure \ref{fig:causal_effect_full}, we present the effect of masking out style-correlated concepts from full images on the model's prediction of that style, compared to two random masking baselines. Targeted concept masking results in the greatest drop in log-probability across all settings, followed by random and then random non-overlapping masking. The difference between concept and random masks are statistically significant in almost all cases, suggesting that the identified concepts are indeed critical to the model's prediction of that specific style. 
\section{Expert Analysis}
In the following, we present two studies we conducted with art historians on the team: an {\em intrinsic} study to analyze the quality of the concepts used by VLMs and an {\em extrinsic} study to assess their alignment with art historians' expertise for the task of style prediction. The extrinsic study used a blind design, where the art historians assessed the degree to which concepts were relevant to style prediction, without knowing which concepts the models predicted as being strongly associated with style. 
%\esc{IMO if we are going to call these user studies, we should more clearly state what the use case is. iirc this is something we discussed in the meetings, but we should be clear here on how we expect the art historians to interact with this system (as users). Or otherwise it's also fine to skip the term ``user study'' and call it an expert evaluation or something like that. }

\subsection{Intrinsic Evaluation: Concept Quality}
\label{sec:concept_assessment}
%\kmnote{As was, this used terms here for the second study and not the first}
%To evaluate the extent to which the extracted concepts 
%KM while both studies do this, the first study is more about concept quality. Let's stick with that wording. The second study is more about alignment. 
%correspond to established domain knowledge in art history,
To evaluate the quality of extracted concepts, 
we conducted a study with six art historians 
%\mbc{was this a blind study in some way to avoid any issues from reviewers about potential subconscious bias from co-authors etc. that we avoid in regular AI papers? this is a different situation because the historians are not really involved in our AI details much but still reviewer might ask; this is why my intro comment might be useful here too i.e. have a discriminative task to check causal/correlation alignment by showing correct vs. incorrect patch-image pairs and asking blindly which ones are better etc. (or patches from baselines) etc.?}
from our research team. The experts were asked to assess the semantic and stylistic coherence of the extracted concepts. Specifically, for each extracted concept, we presented 24 image patches representative of that concept and instructed them to (1) provide up to three textual labels for what the patches have in common, which we use to evaluate our automatic labeling methods, and (2) rate the degree to which the images collectively represent a single, coherent art-historical concept on a 5-point Likert scale. Each concept was evaluated by three art historians. To avoid potential bias among the experts, we do not present them with the automatically generated textual labels. Instead, we exclusively employ the labels they provide to evaluate the quality of our generation methods.
%\kmnote{I think we should cut the following sentence. It's not critical. Someone may ask us to compute interannotator agreement separately for the batches. }
%For convenience, we split the 128 concepts into two batches, and each batch was evaluated by a different group of three art historians.
%Moreover, in a follow-up qualitative analysis, 
We also compare the concepts extracted at the image level (32 concepts from prior work) and the patch level (128 concepts from our work) in terms of their thematic coverage; one art historian classified each image and patch level concept using the art historians' textual labels as guidance into one of three categories: reflecting content, form, style, or reflecting no coherent theme.

\begin{table}[h]
\centering
\begin{tabular}{lccccc}
\toprule
\textbf{Concept Coherence} & 1 & 2 & 3 & 4 & 5 \\
\midrule
\textbf{Number of Concepts} & 14 & 21 & 33 & 43 & 17 \\
\bottomrule
\end{tabular}
\caption{Distribution of scores given by art historians for the coherence of the extracted concepts}
\label{tab:concept_coherence}
\end{table}

\begin{figure}
    \centering
    \includegraphics[width=0.8\linewidth]{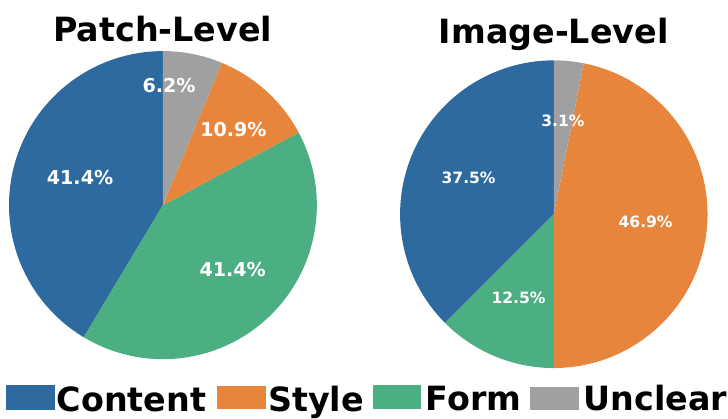}
    \caption{Percentage of concepts for each theme presented for image-level (prior work) and patch-level (our work)}
    \label{fig:content_form_style}
\end{figure}

\subsubsection{Results}

\begin{figure}
    \centering
    \includegraphics[width=0.99\linewidth]{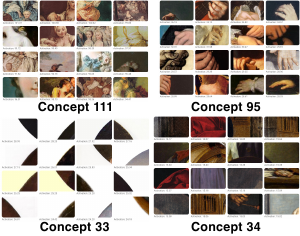}
    \caption{Example concepts discussed by art historians}
    \label{fig:example_concepts}
\end{figure}

We collect a majority label for each concept and report our results accordingly. To measure inter-annotator agreement, we compute Krippendorff's alpha, which yields 0.52, indicating moderate agreement on concept coherence scores. Table \ref{tab:concept_coherence} shows the distribution of coherence scores given by art historians. Of the 128 concepts assessed, 93 (73\%) received a majority score of 3 or higher, indicating the majority of concepts do represent a coherent concept from the art-historians' perspective. 
%Table \ref{table:arthistorians-labels-given-for-top-concepts} shows the top three ranked concepts, and the labels that were given by the art historians. \kmnote{Marvin - we did not discuss this, but here is where I think you should add the visual concepts along with the labels and indicate for each whether content, form or style. Hopefully you have different types.} Some of these concepts, like Concept 1, reflect color, while others, like Concept 19, reflect geometric shapes. 
\paragraph{What themes do concepts cover?} In Figure~\ref{fig:content_form_style}, we show the distribution of concepts across content, form, style, or no coherent theme for image-level and patch-level concepts. For image-level concepts, we observe that the largest category was stylistic theme. This theme only indicates that a concept was characteristic of a particular style, but provides  no explanation of how different content and form aspects contribute to the model's predictions. In contrast, relatively few patch-level concepts seem primarily bound by a specific style; one exception is concept 111, shown in Figure \ref{fig:example_concepts}.

A vast majority instead seem primarily united by form (e.g., concept 33) or by content (e.g., concept 95). Most often the concepts are united by some combination of form, content, and style, with each present to varying degrees (e.g., in concept 111, all of the top-activating patches are “Rococo” but in addition most convey a certain kind of content—here mainly details of women). For only a few concepts, it is unclear, from a human perspective, how they are united—whether by style, form, or content (e.g., concept 34). 
\begin{comment}
\begin{figure}
    \centering
    \includegraphics[width=0.7\columnwidth]{figures/concept_coherence.pdf}
    \caption{Distribution of scores given by art historians for the coherence of the extracted concepts}
    \label{fig:concept_coherence_score_dist}
\end{figure}
\end{comment}

In art history, style categorizes artworks historically by their intrinsic qualities \cite{schapiro1953style}, encompassing both formal aspects (contour, scale, color, tone, texture, etc.) and content (what an artwork represents, means, or communicates) \cite{goodman1975status}.
Though style is a messy concept whose value for art history is often questioned \cite{joyeuxprunel2024style, elsner2003style}, it remains central to art historical research and teaching.

%MA: Shortened above
% It is worth noting that style in art history is a way of categorizing artworks historically based on their intrinsic qualities \cite{schapiro1953style}, and includes not only aspects of the form (contour, scale, color, tone, texture, etc.) of artworks but also their content (what an artwork represents, means, or otherwise communicates) \cite{goodman1975status}. 
% %\kmnote{I would like to cut this sentence.}
% %These definitions are all contested in art history, but for clarity, we use the terms as described here. 
% While style is a messy concept whose usefulness for art history has often been called into question \cite{joyeuxprunel2024style, elsner2003style}, it remains a primary feature of art historical research and teaching.

% We highlight that style in art history is a way of categorizing artworks historically based on their intrinsic qualities, and which includes not only aspects of the artworks’ form (contour, scale, color, tone, texture, etc.) but also their content (what the artwork represents, means, or otherwise communicates). The definitions of style, form, and content are all contested within art history, but for the sake of clarity, here we use the terms as just described. \kmnote{I feel this could be more carefully written. As is, it will cause concern among CS reviewers. Can we scale back? } \esc{Maybe the art historians have some canonical citation for this point? Like some art history position paper that argues that these are what should be included in ``style'' even if that's contested by others}

\paragraph{Alignment between expert and automatic labels} 
Given the expert labels collected from this study, we analyze the quality of our labeling methods using \texttt{GPT5.4}-as-a-Judge. Our analysis showed that the summary based method, where an LLM generates a label from concatenated descriptions of the top activating images, was quite effective in generating concept labels that compared well with expert labels (92\% win rate compared to other models) compared with a baseline (win rate of 21\%) and two other methods, CLIP image similarity without and with negative samples (win rates of 39\% and 49\%, respectively). Detailed results can be found in Appendix \ref{sec:appendix_concept_labeling}. Although this evaluation is not conclusive, it provides preliminary evidence that generating high‑quality captions for visual patch concepts is feasible. 
\begin{figure}
    \centering
    \includegraphics[width=1\linewidth]{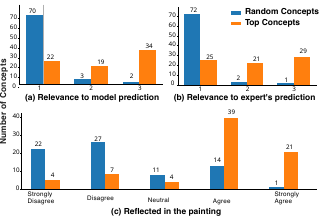}
    \caption{Expert Analysis 2: Concepts' ranking on a scale from 1 to 3 based on their relevance to the model prediction (a) and the art historian's prediction (b) as well as their scores for being 
    reflected in
 %   relevant to 
    paintings (c).}
    \label{fig:user-study-2}
\end{figure}
%\subsection{Alignment of Concepts for Style Prediction}
\subsection{Extrinsic Evaluation: Style Prediction}
\label{sec:art_interpretability}
In this analysis by art historians, we evaluate the plausibility of concepts for style prediction. To this end, we designed an experiment in which we presented six art historians with a piece of artwork together with the model’s style prediction and three associated concepts, shown in  Figure~\ref{fig:case2}. For each image, up to two of these concepts were selected from the top associated concepts identified in the image, while the remaining concepts were randomly chosen from non-activated concepts for calibration. The art historians were first asked to infer the artwork’s artistic style, then to evaluate the extent to which each concept is reflected
in the artwork, and finally to rate the relevance of each concept to their own predicted style, as well as to the model’s predicted style when it differed from their own. If annotators assign high ratings to the top associated concepts but low ratings to the control concepts, we take this as evidence that the concepts are plausibly associated with the predicted style and aligned with art historians' domain knowledge. We assembled 50 cases for the study, selecting ten from each of the examined art styles; for each style, we chose cases where the model correctly predicted seven cases and failed to predict three to match model performance on all images. 

\subsubsection{Results}
From the six annotations, we aggregate the majority label from the art historian's prediction of style, as well as %for 
their ranking of the concept's relevancy.

\paragraph{Agreement on Style Prediction} We measure stylistic consensus among art historians using Fleiss’ kappa over 50 cases, and assess pairwise agreement between WikiArt and each historian with Cohen’s kappa. The six annotators achieve a Fleiss’ kappa of 0.52, indicating moderate agreement, and the mean pairwise Cohen’s kappa between WikiArt and the historians is 0.55, also moderate agreement.

%KM Removed for now. 
%\manote{The following might backfire on the reliability of our insights:}Interestingly, The art historians agreed only in 53\% of the cases with 
%KM I don't think we need to say this. It is the ground-truth we and others use.
%what we consider 
%the ground-truth style in the dataset.
\paragraph{Concept Relevance} Figure \ref{fig:user-study-2} shows the distribution of majority scores achieved by the most associated concepts to the predicted style compared to the random ones. 
%As for the relevance of these concepts to the painting (Figure \ref{fig:user-study-2}.c), 
Only 17\% (13 out of 75) of the top associated concepts were marked as not reflected in the painting compared to 70\% of the random ones (Figure \ref{fig:user-study-2}.c). This result demonstrates that our approach extracts concepts that art historians consider reflected in the artwork. 
%Considering concepts' relevance to the model's prediction (Figure \ref{fig:user-study-2}.b), 
Out of 75 concepts that %were considered 
%KM making explicit the difference between model relevance and art historian relevance
the model determined were associated with the style prediction, 29\% (22 concepts) received a score of one (not relevant) by the art historians (Figure \ref{fig:user-study-2}.b). When the art historians' judge relevance to their own prediction of style, the percentage  of irrelevant concepts increases to 33\%  (Figure \ref{fig:user-study-2}.a). Compared to randomly selected concepts, which were mostly considered irrelevant to the prediction (93\% and 96\%), we conclude that there is substantial alignment between the model's concepts and art historians' domain knowledge.

\subsection{Qualitative Analysis of Concepts}
\label{sec:qualitative_analysis}

% concept multiplicity -- individual concepts have several unifying threads that allow the model to flexibly compose them in predicting different styles
% concept bias -- content over-represented in the images of a particular period but not canonically understood as indicate of a style; spurious correlations

% say something about human vision vs machine vision
The art historians closely examined cases where the relevance of the model's top activating concept was not immediately obvious.  These cases, which include both correct and incorrect predictions, shed light on two differences between expert and machine style recognition: \textit{concept multiplicity} and \textit{content bias}.

\begin{figure}
    \centering
    \includegraphics[width=\linewidth]{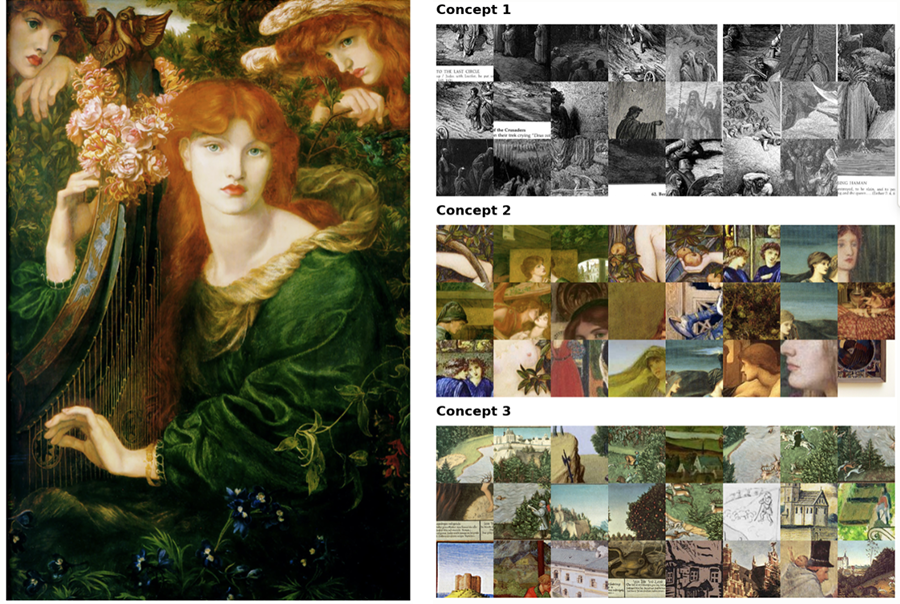}
    \caption{Example artwork with top associated concepts 1 \& 2 and a random concept 3}
    \label{fig:case2}
\end{figure}

\textit{Concept multiplicity.} Despite the coherence of our concepts, many exhibit multiple unifying threads.  This is distinct from polysemanticity observed in other interpretability work \cite{elhage2022toy} -- in the cases we present here, all of the patches associated with a concept exhibit a conjunction of form and content elements.  The model flexibly exploits such concepts in predicting different styles, contextually isolating what is most relevant to a given image.  For example, in Figure \ref{fig:case2}, the model relies on concept 1 to correctly predict Romanticism for Dante Gabriel Rossetti’s \textit{The Garland} (1873).  When considering concept 1 in isolation,  it was not immediately clear to the art historians why concept 1 was relevant.  
Concept 1 consists primarily of Romanticist black-and-white print illustrations, specifically those by Gustave Doré (1832–83) for Dante’s Divine Comedy. Obvious differences of form (color, medium) as well as differences in content between the concept and the image likely contributed to their judgment of concept 1 as irrelevant to prediction of Romanticism. 
However, they could make sense of concept 1's inclusion within the context of its other activated concepts as in aggregate they are clearly linked to Romanticism.  

In Figure \ref{fig:appendix_user_study_2_UI_1} (see Appendix), the model relies again on concept 1, this time when incorrectly predicting Baroque for a painting from the Northern Renaissance.  The use of the same concept in representing diverse imagery suggests the model makes connections between different aspects of the same concept and different styles.  
%speaks to its ability to disentangle relevant elements of form, content and style. and 
This highlights the presence of compositional neural circuitry driving both correct and incorrect predictions. While correct predictions are explained by this compositionality, incorrect decisions require further investigation.

\textit{Content bias.} A common source of misalignment between the art historians and the model was driven by the model's over-dependence on content.  In such cases, the model relies on content commonly represented in the imagery of a particular style but not canonically understood as prototypical of it.  For example, the model associated a concept corresponding to trees and forests with Romanticism much more strongly than the art historians.  While nature plays an important role in much Romanticist art, when presented with that concept in isolation, our art historians did not make this connection.  This suggests that content may sometimes play 
%a more exaggerated role 
more of a role
in model style classification than it does for art historians. This is potentially driven by phenomena similar to those underlying object bias in VLMs and merits more study \cite{wan2024contrastive}.

\section{Discussion and Conclusion}

We present a method to extract interpretable visual concepts from latent representations in models and  
%their concept-style associations. We 
and demonstrate the usefulness of our approach 
%on \texttt{Qwen3} and \texttt{Llava-1.5} 
using the WikiArt dataset, finding that style-specific visual concepts emerge throughout the model’s layers and
can partially explain observed differences in model performance {\bf (RQ1)}. Our  analysis shows that activated concepts are strongly correlated with the model's predictions (95\%) accuracy in the final layers of \texttt{Qwen3} and causal analysis shows that masking out the top associated concepts results in a significantly greater reduction in accuracy of style prediction than when masking out random concepts {\bf (RQ2)}. 

Our expert analysis with art historians yields nuanced answers to {\bf RQ3:  Do visual concepts identified by VLMs align with how art historians see style?} On the one hand,  %in our art historian study,
the art historians note that in 73\% of cases they analyzed 
%for style prediction,
%of a given artwork,  
the top associated concepts are 
%reflected in the artwork and are 
relevant to prediction of style. This suggests that, in the majority of cases, the model does align with how art historians see style. In the cases of misalignment, however, interesting observations emerge. 
%The model is better able than humans to identify multiple connections between concepts and artwork.
The model relies on multiple and nonobvious (to art historians) connections between concepts and artwork.
%This happens when the model relies on multiple aspects of a concept, 
%beyond 
For example, using multiple concept aspects, it overcomes the lack of similarity in color and form between the concept and the image to make other connections with style (Romanticism, in case 1, Rosetti). 
%It also can make use of the same concept for classifying images to quite different styles, likely deriving in part from drawing on different features within the cluster of patches representing a concept. 
This exploitation of concept multiplicity was not immediately obvious to art historians.
%until they examined a much larger set of associated concepts.
%associated with specific instances of style prediction. 
This tendency in models thus suggests that there are cases where the model's vision operates differently than art historian's view, relying on different patterns to make it predictions. 

\section{Limitations and Future Work}

%First, the fixed 4×4 image patching occasionally caused features to be cut off. 

%KM cut to save space

%Given occasional errors resulting from using  fixed 4x4 image patching (e.g., in some cases features are cut off), we suggest that future work carry out a thorough comparison with alternative patching strategies such as segmentation or native VLM tokenization.   These approaches introduce additional dependencies (e.g., segmentation model, dynamic tokenization for varying image sizes and VLMs) so it is unclear if they would improve on our method. 
%Second, 
Future work could  explore additional expert analysis studies to arrive at a deeper interpretation of differences between  the model and art historian.  
%For this paper, in post-analysis, they found it helpful 
% when they were given additional details for some cases such as the image motif and additional relevant patch concepts and this insight 
 For example, a larger study involving analysis of additional relevant patch concepts could provide new insights.
 %, which aligns with the holistic nature of style prediction. 
Furthermore, due to the two-step image-patch concept mapping method, occasionally patch concepts presented to the experts were not directly present in the image.
%---rather, patch concepts were  strongly associated with the broader image motif---
%leading to incoherence. 
Future work could explore reliance on the causal analysis to identify important patches, although this is computationally intensive. Finally, a future larger-scale study of automatic labeling incorporating human evaluation would enable a more rigorous investigation and validation of possible approaches.
%One option is to adopt the causal analysis approach to identify important patches, though this is more intensive, requiring numerous inferences per image and accounting for combined patch effects.
%Future work adopting this approach may focus on aligning the presentation of results with the target domain.

\section*{Acknowledgements}

This research was partially funded through support from Columbia Data Science Institute's seed funding program. 

\clearpage

\bibliographystyle{acl_natbib}   % or whatever style TACL requires
\bibliography{tacl2021}

\clearpage
\appendix
\section*{Appendix (Category 1)}  % unnumbered header
\addcontentsline{toc}{part}{Category 1}

\section{Concept Decomposition Details}

\subsection{Prompt}

We include the prompt used for classifying art style below. This prompt was provided along with an image as input to the VLM to generate the latent representations for concept decomposition. For the causal analysis, we modify the prompt slightly, formulating it as a multiple choice question (A, B, C, ...) to better measure the probability of the VLM predicting each style.

\begin{tcolorbox}[title=Prompt: Style Classification, fontupper=\small\ttfamily]
Identify the style of this \texttt{[artwork $|$ architecture]} as one of the following: \{STYLE\_1\}, \{STYLE\_2\}, \{STYLE\_3\}, \{STYLE\_4\}, or \{STYLE\_5\}. You MUST output exactly one of these style names first, then begin the rest of your response.
\end{tcolorbox}

\subsection{Metrics}

\textbf{Effective Labels} (Eq. \ref{eq:efflabels}) ranges from $[1-5]$ and captures how ``pure'' (on average) concepts $c$ are with respect to the style labels $y$ (ground-truth or predicted) of its activating images. A value of 1 indicates all images in the concept share the same style; $5$ indicates uniform mixing.
\begin{equation}
    \text{EffLabels}(c) = \exp\!\Bigl(-\textstyle\sum_y\, p(y \mid c)\log p(y \mid c)\Bigr) \vphantom{\sum_y}
    \label{eq:efflabels}
\end{equation}
\textbf{Sparsity} (Eq. \ref{eq:sparsity}) measures the average sparsity of concept activation vectors (one per image) using the Hoyer measure, ranging from $[0 - 1]$. A value of 0 indicates all concepts contribute equally; 1 indicates a single concept dominates.
\begin{equation}
    H(\mathbf{v}_i) = \frac{\sqrt{K} - \|\mathbf{v}_i\|_1 / \|\mathbf{v}_i\|_2}{\sqrt{K} - 1}
    \label{eq:sparsity}
\end{equation}
\textbf{Reconstruction error} (Eq. \ref{eq:recerr}) is the mean per-sample relative $\ell_2$ error between the original representation $\mathbf{Z}$ and its reconstruction $\mathbf{UV}$. Lower values indicate the concept dictionary more faithfully recovers the original representations.
\begin{equation}
    \text{Recon. Error} = \frac{1}{n}\sum_i \frac{\|\mathbf{z}_i - \mathbf{U}\mathbf{v}_i\|^2}{\|\mathbf{z}_i\|^2}
    \label{eq:recerr}
\end{equation}

\section{Concept Labeling Methods}
\label{sec:appendix_concept_labeling_methods}

As part of our expert analysis, we collect reference labels for each visual concept annotated by three art historians. For those adapting our study to new domains, we present three methods of automatically generating such labels

\begin{enumerate}
    \item \textsc{Baseline}: Following \citet{parekh2024concept}, we apply the Logit Lens technique---projecting concept vectors through the LM head to rank tokens by logit score. After filtering stopwords and non-English words, the top 10 tokens are concatenated as a single string label.

    \item \textsc{Summary}: The top $m=20$ images ranked by activation are selected. For each image, we prompt \texttt{GPT-5.2} to generate a description focusing on stylistic and art-historical properties. These descriptions are concatenated and passed to the model to generate a single label capturing the dominant pattern(s) that describes the concept. We note that this method is closely aligned with existing work on LLM interpretability \citep{bills2023language}.
    
    \item \textsc{Image-Sim}: We concatenate the $m$ descriptions from \textsc{Summary} and prompt \texttt{GPT-5.2} to generate up to five candidate labels. The label with the highest average \texttt{CLIP} cosine similarity score to the $m$ images is selected.
    
    \item \textsc{Contrastive}: \textsc{Image-Sim} with negative sampling. The \textsc{Image-Sim} procedure is repeated, but label selection incorporates a negative term computed over $m$ images sampled from the 5 nearest concepts (by cosine similarity). The label maximizing this new score is selected.
    
\end{enumerate}

\begin{tcolorbox}[
    breakable,
    skin=enhanced jigsaw,
    title=Prompt: Image Description, 
    fontupper=\small\ttfamily
]
You are given an image patch taken from an artwork. Your task is to generate a paragraph-length description of the image patch. This description should include the content of the image (i.e., what things are being portrayed) and visual properties corresponding to art style (see below for more details). For visual properties, consider some of the following. Color: What kinds of colors are used in the work? How would you characterize the color palette and tone of the colors (i.e. how light or dark)? Texture: Are there different textures in the work? Is the texture smooth, rough? If there are brushstrokes, are they visible, subtle, loose, flowing, tight, controlled, etc.? Composition: How are figures or objects and space arranged? Are there interaction between these elements? Is it dynamic or still? What about the proportions of the individual parts of the image? What is the relation of the image to the entire artwork composition? Light and Shadow: Is light directed from a particular angle? Are there light or dark parts of the image? What about shadows? Pose: If the work has figures, how are their bodies posed (rest, graceful, relaxed, tense, etc.)?
[FORMATTING INSTRUCTIONS ...]
%\\ \\
%DO NOT elaborate on external factors such as the artist, artwork title, specific art style, or other context about the artwork. Focus on the elements reflected in the image patch itself. Output your response as one long paragraph beginning with "Description: ..."
\end{tcolorbox}

\section{Causal Analysis Methods}

\subsection{Patch-Level Causal Analysis}
To supplement the linear probe analysis, we verify that patch-level concepts causally influence the model's style prediction on image patches. Specifically, on a held-out set of image patches where the prediction was correct, we measure $\log p(s \mid \mathbf{x})$, the baseline log-probability of each style $s$ being generated using teacher forcing over the first two tokens (two to distinguish unique style names). Then, for each of the top three activated concepts, we directly modify the model's hidden state $\mathbf{h_{L}}$ on that patch example by subtracting out the concept direction $\mathbf{v_{i}}$
\begin{equation}
    \tilde{\mathbf{h_{L}}} = \mathbf{h_{L}} - (a_i\mathbf{v}_i)
    \label{eq:intervention}
\end{equation}
where $L$ is the layer and $a_{i}$ is the concept's activation. This effectively `zeros' out the concept from the latent representation. This modified hidden state is propagated through the VLM, the log-probabilities are re-computed, and the \textbf{causal effect} of the intervention $i$ on a style is
\begin{equation}
    \Delta_i(s) = \log p(s \mid \tilde{\mathbf{x}}_i) - \log p(s \mid \mathbf{x}).
    \label{eq:causal_effect}
\end{equation}
Here, $\tilde{\mathbf{x}}_i$ denotes that the same input is provided but the hidden state was intervened on. 
%We normalize log-probabilities over the set of styles before taking the difference to ensure deltas reflect redistribution among styles. 
To control for random effects, the causal effect of each intervention is calibrated against that of 10 random directions of equal magnitude.

\subsection{Full Image Concept Masking}
\label{sec:appendix_full_image_concept_masking}
\begin{figure}[h!]
    \centering
    \includegraphics[width=0.97\columnwidth]{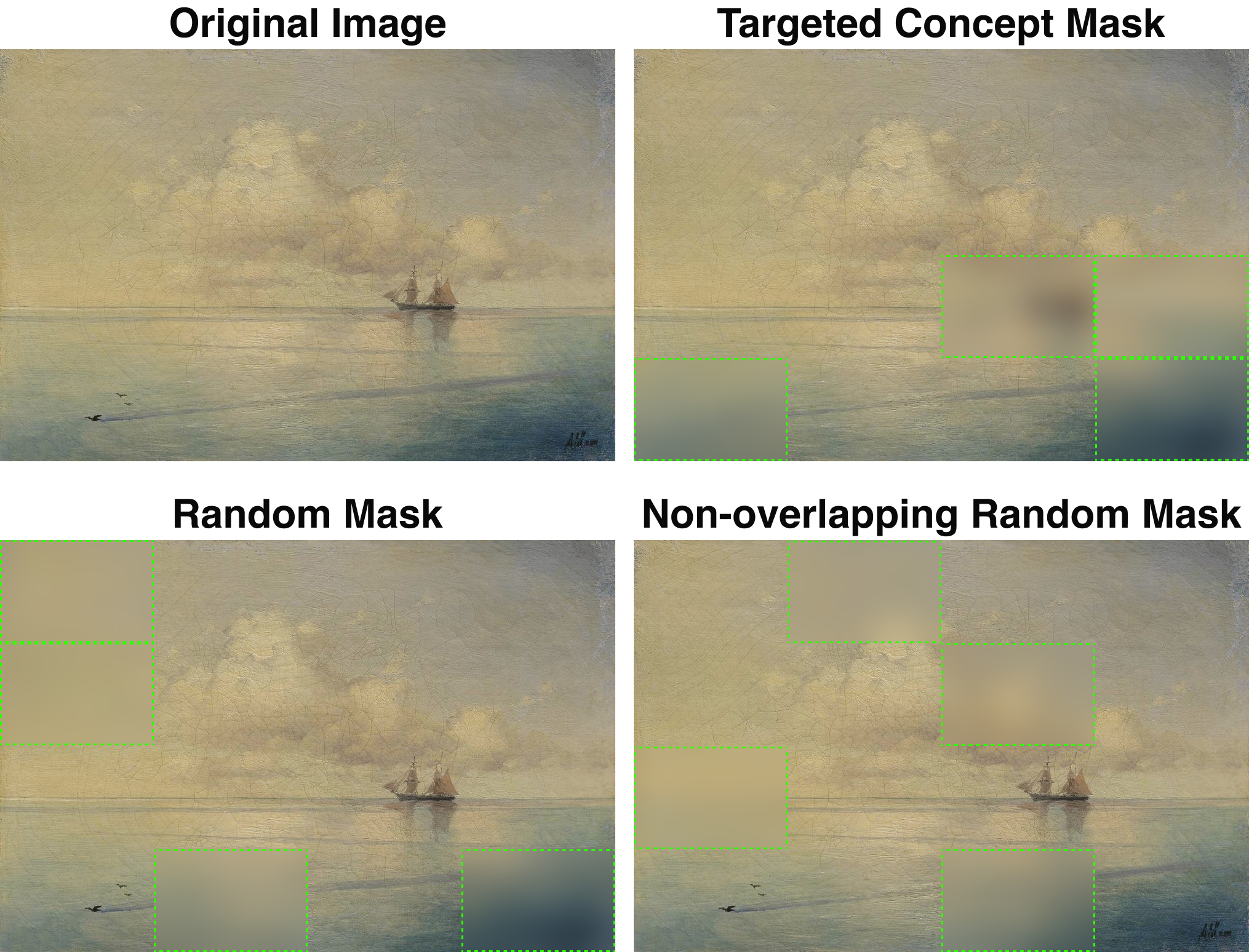}
    \caption{An example of our causal masking settings. }
    \label{fig:appendix_causal_mask}
\end{figure}

\begin{figure}[h!]
    \centering
    \includegraphics[width=0.97\columnwidth]{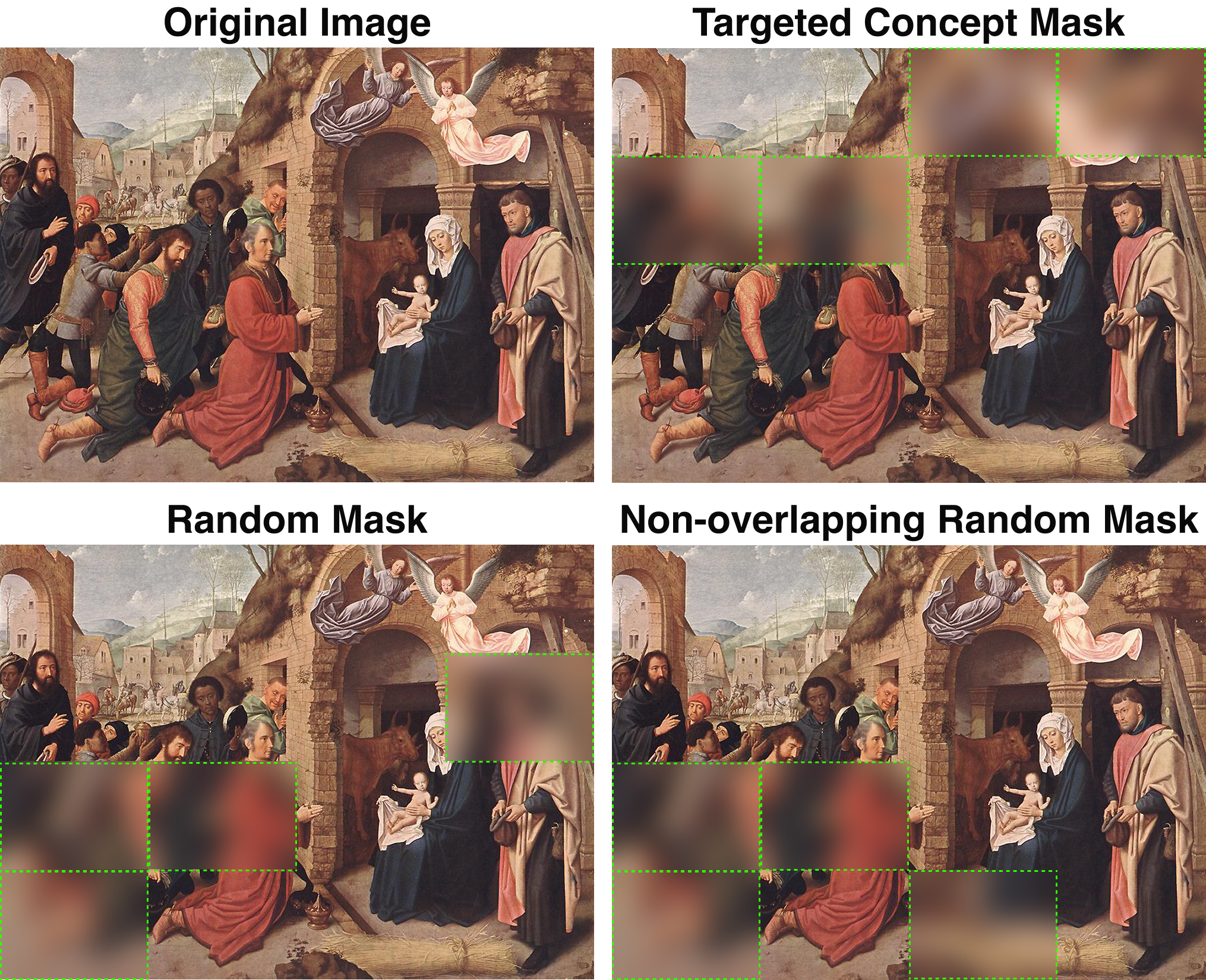}
    \caption{An example of our causal masking settings. }
    \label{fig:appendix_causal_mask_2}
\end{figure}

Figure \ref{fig:appendix_causal_mask} shows an example from our causal masking experiment. Patches corresponding to the ship, blue water, and birds are identified as causally relevant to ``Romanticism'' and blurred in the targeted mask. The random mask may overlap with these regions by chance (e.g., the blue water), while the non-overlapping random mask explicitly excludes them. An additional example for a ``Renaissance'' example is shown in Figure \ref{fig:appendix_causal_mask_2}.

\clearpage
\begin{figure*}[t!]
    \section{Annotation Guidelines}
    \label{sec:appendix_annotation_guidelines}
    \begin{minipage}{\textwidth}
        \centering
        \includegraphics[width=\textwidth]{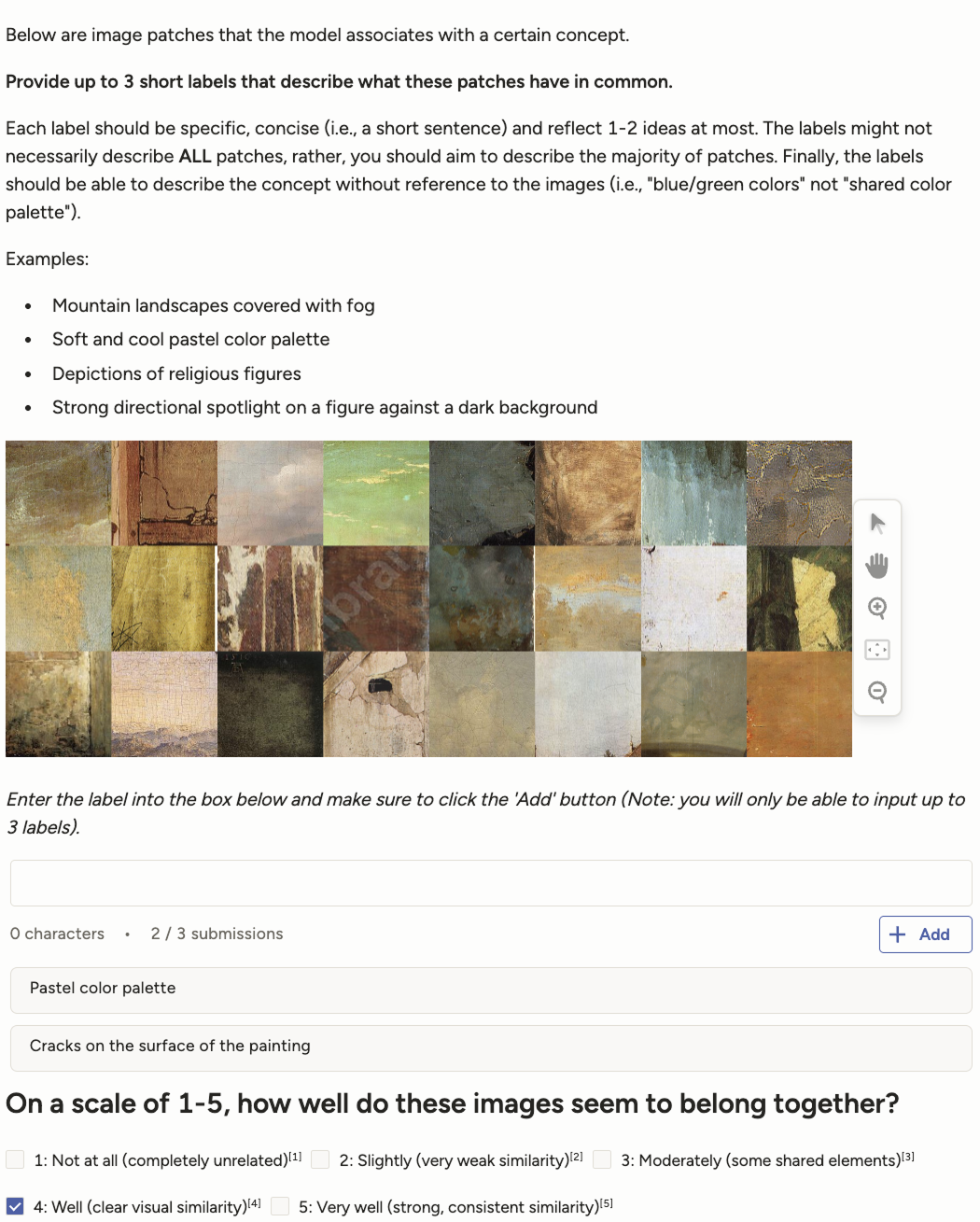}
        \caption{User interface for Study 1. Annotators were first presented with labeling instructions and examples, then shown a 3×8 grid of top-activating image patches (each taken from a distinct image) representing a visual concept and asked to provide up to three short descriptive labels, followed by a coherence rating.}
        \label{fig:appendix_user_study_1_UI}
        \vspace{6em}
    \end{minipage}
\end{figure*}
\clearpage
\begin{figure*}[t!]
    \begin{minipage}{\textwidth}
        \centering
        \includegraphics[width=0.99\textwidth]{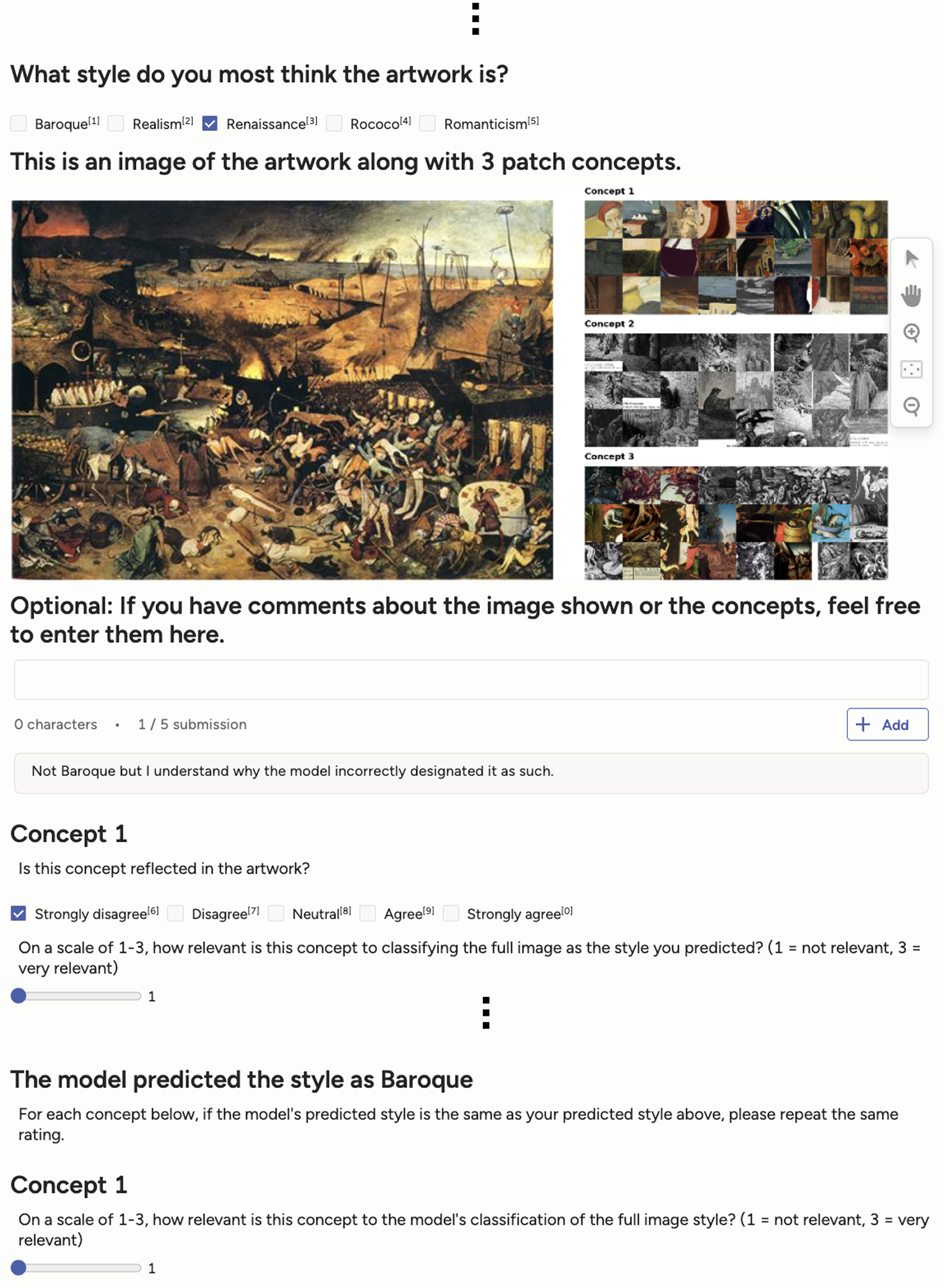}
        \caption{User interface for Study 2. Participants were first shown the image alone (not pictured) and asked to predict its style; this answer could not be revised once submitted. They were then presented with three concepts (a mix of random and relevant) and asked whether each was reflected in the image and how relevant it was to their prediction. They additionally could enter an optional text comment. Finally, the model's prediction was revealed and the same questions were repeated. Once seeing the model's prediction, previous answers could not be revised. This image is Pieter Bruegel the Elder's The Triumph of Death (labeled as Northern Renaissance in WikiArt).}
        \label{fig:appendix_user_study_2_UI_1}
    \end{minipage}
\end{figure*}
\clearpage
\begin{figure*}[t]
    \section{Additional Concept Visualizations}
    \label{sec:appendix_additional_concept_examples}
    \begin{minipage}{\textwidth}
        \centering
        \includegraphics[width=\textwidth]{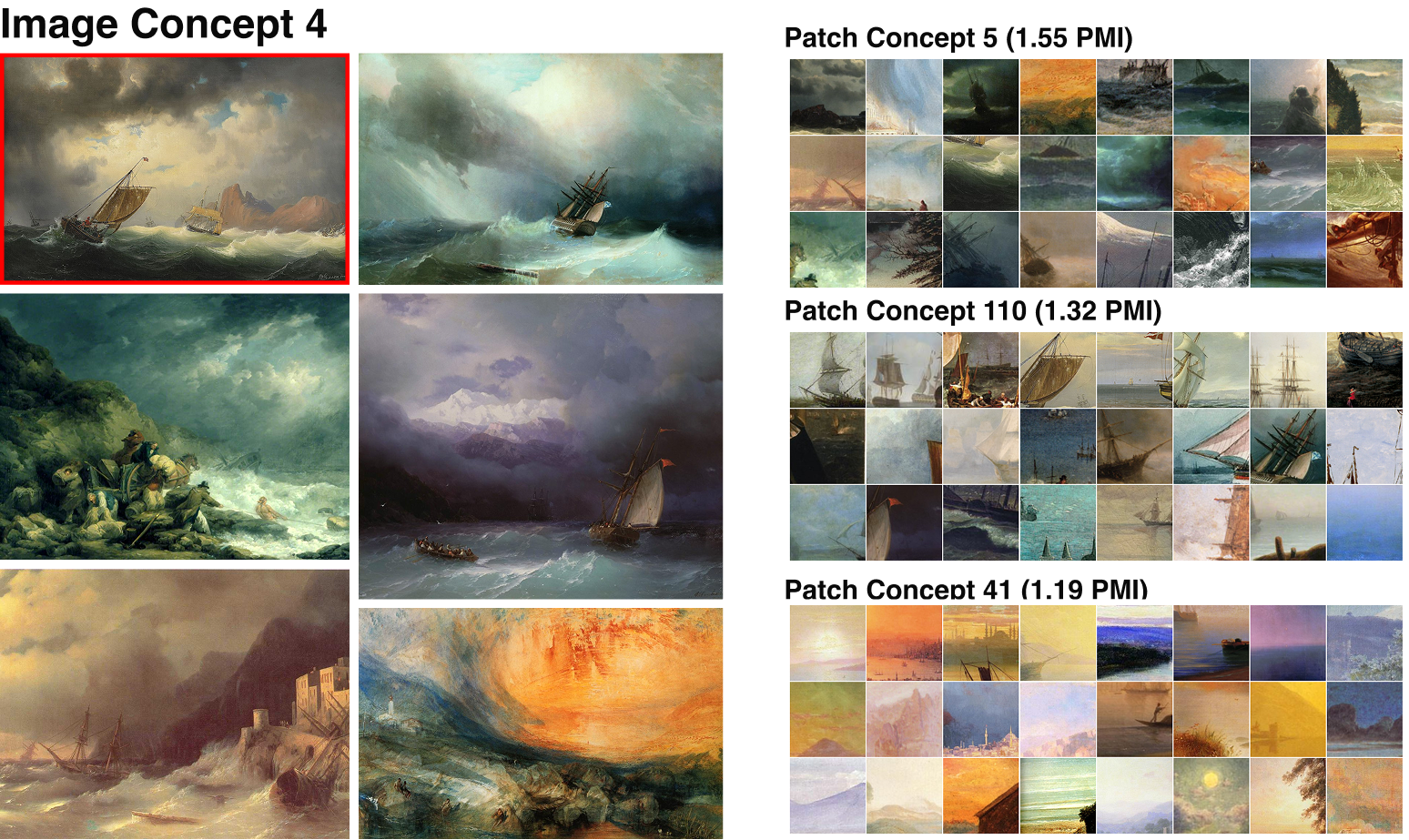}
        \caption{An example of concept mapping inputs/outputs. \textbf{Left:} The input image (red outline), Marus Larson's \textit{Skepp på stormigt hav}, is mapped to image concept 4. \textbf{Right:} The most associated patch concepts for image concept 4 are displayed on the right, ordered by PMI score.}
        \label{fig:appendix_image_concept_1}
        \vspace{1em}
    \end{minipage}
    \vspace{1em}
    \begin{minipage}{\textwidth}
        \centering
        \includegraphics[width=\textwidth]{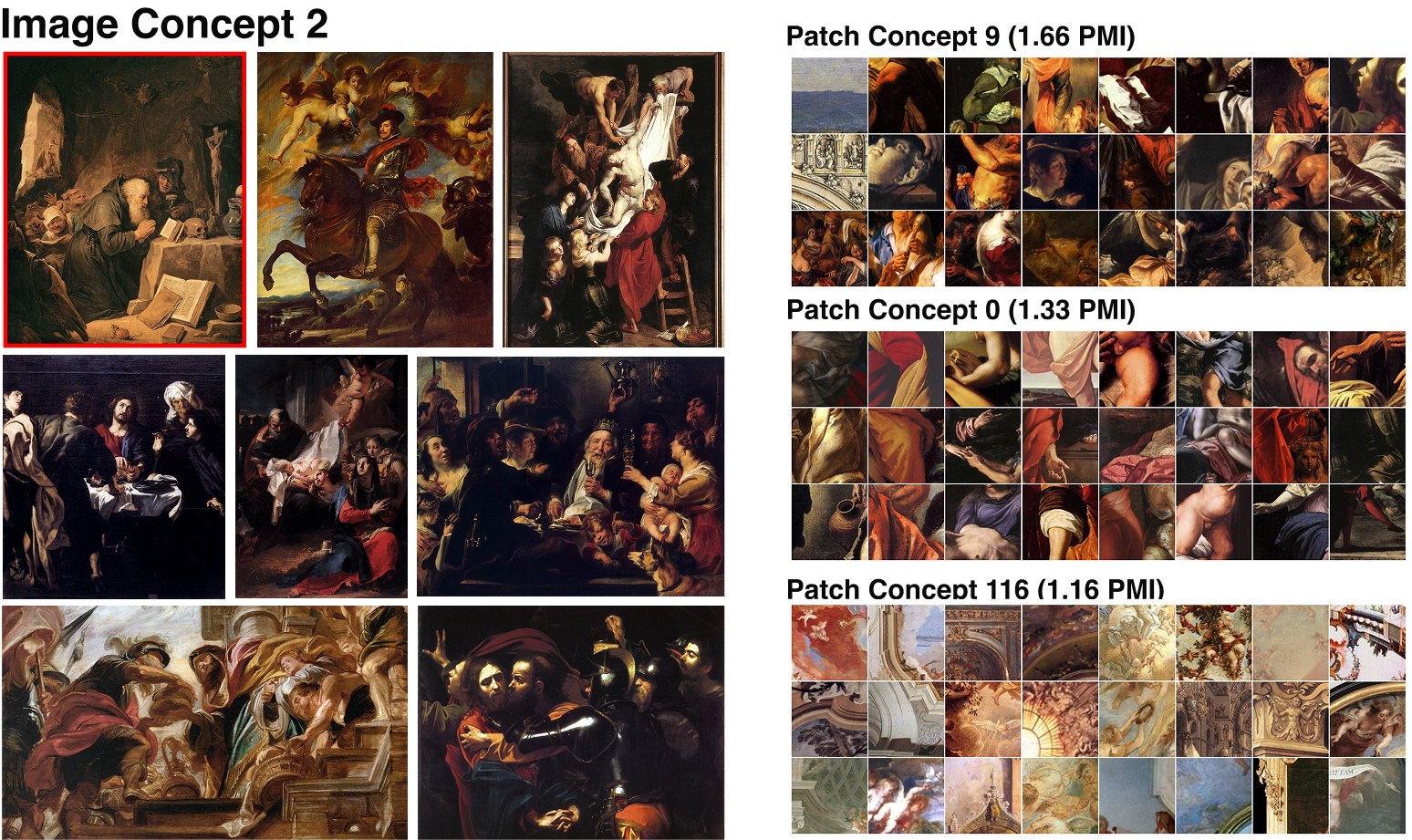}
        \caption{An example of concept mapping inputs/outputs. \textbf{Left:} The input image (red outline), David Teniers II's \textit{The Temptation of St. Anthony}, is mapped to image concept 2. \textbf{Right:} The most associated patch concepts for image concept 2 are displayed on the right, ordered by PMI score. Note how here, the third patch concept is not reflected in the image.}
        \label{fig:appendix_image_concept_2}
    \end{minipage}
\end{figure*}
\clearpage
\section*{Appendix (Category 2)}  % unnumbered header
\addcontentsline{toc}{part}{Category 2}

\section{Model Benchmarking}
\label{sec:appendix_2_model_benchmarking}

We include a per-style breakdown of VLM accuracies for art style prediction in Table \ref{tab:appendix_per_class_accuracy}. In the case of \texttt{LLaVA-1.5}, high accuracies can reflect a strong prediction bias towards certain styles.

\label{sec:appendix_model_benchmarking}
\begin{table}[h!]
\centering
\small
\definecolor{sectiongray}{gray}{0.92}
\setlength{\tabcolsep}{3.5pt}
\begin{tabular}{p{1.75cm}cccc}
\toprule
\textbf{Style} & \textbf{GPT5} & \textbf{Qwen3} & \textbf{Molmo2} & \textbf{LLaVA1.5} \\
\midrule
\rowcolor{sectiongray}
\multicolumn{5}{l}{\textbf{WikiArt (Early Mod.)}} \\
Baroque & \underline{.73} & \textbf{.86} & .52 & .80 \\
N. Renaiss. & \underline{.88} & \textbf{.91} & .74 & .14 \\
Realism & \textbf{.76} & \underline{.58} & .46 & .05 \\
Rococo & .48 & \textbf{.51} & \underline{.37} & .12 \\
Romanticism & .61 & \underline{.67} & .44 & \textbf{.82} \\
\textit{Overall} & \underline{.69} & \textbf{.71} & .51 & .39 \\
\midrule
\rowcolor{sectiongray}
\multicolumn{5}{l}{\textbf{WikiArt (Modern)}} \\
Abstr. Expr. & .71 & \underline{.73} & .90 & \textbf{.99} \\
Color Field & \underline{.49} & .41 & \textbf{.67} & .09 \\
Cubism & \textbf{.75} & \underline{.74} & .28 & .04 \\
Fauvism & \underline{.87} & \textbf{.96} & .82 & .10 \\
Minimalism & \textbf{.89} & \underline{.83} & .57 & .24 \\
\textit{Overall} & \textbf{.74} & \underline{.73} & .65 & .29 \\
\midrule
\rowcolor{sectiongray}
\multicolumn{5}{l}{\textbf{Architecture}} \\
ArtNouveau & \underline{.75} & \textbf{.89} & .39 & .76 \\
Baroque & \textbf{.92} & \underline{.89} & .70 & .58 \\
Byzantine & \underline{.86} & .78 & \textbf{.85} & \textbf{.89} \\
Gothic & .97 & \textbf{.98} & \underline{.98} & .58 \\
Romanesque & \textbf{.85} & \underline{.79} & .21 & .78 \\
\textit{Overall} & \textbf{.88} & \underline{.84} & .64 & .72 \\
\midrule
\rowcolor{sectiongray}
\multicolumn{5}{l}{\textbf{WikiArt (Control)}} \\
\textit{Overall} & \underline{.90} & \textbf{.93} & .82 & .78 \\
\bottomrule
\end{tabular}
\caption{The accuracy per style class across models and datasets. Best results in bold, second best underlined. \textit{Overall} reports accuracy over the entire dataset.}
\label{tab:appendix_per_class_accuracy}
\end{table}

\begin{comment}
\begin{figure}[h]
    \centering
    \includegraphics[width=0.96\columnwidth]{./figures/architecture_per_class.png}
    \caption{Accuracy across styles for the Architecture dataset.}
\label{fig:appendix_architecture_per_class}
\end{figure}
\begin{figure}[h!]
    \centering
\includegraphics[width=1\columnwidth]{./figures/wikiart_control_per_class.png}
    \caption{Per-class accuracy for the Wikiart (control) dataset across VLMs.}
\label{fig:appendix_wikiart_control_per_class}
\end{figure}
\begin{figure}[h!]
    \centering
\includegraphics[width=0.96\columnwidth]{./figures/wikiart_early_modern_per_class.png}
    \caption{Accuracy across styles for the Wikiart (early modern) dataset.}
\label{fig:appendix_wikiart_early_modern_per_class}
\end{figure}
\begin{figure}[h!]
    \centering
    \includegraphics[width=0.96\columnwidth]{./figures/wikiart_modern_per_class.png}
    \caption{Accuracy across styles for the Wikiart (modern) dataset.}
\label{fig:appendix_wikiart_modern_per_class}
\end{figure}
\end{comment}
\section{Concept Decomposition Details}
\subsection{Parameters}
\label{sec:appendix_parameters}
Table \ref{tab:appendix_concept_decomposition_metrics} shows the effect of layer and number of concepts \textbf{K} on our metrics.
\begin{table}[h!]
\centering
\small
\definecolor{sectiongray}{gray}{0.92}
\setlength{\tabcolsep}{6pt}
\begin{tabular}{ccccccc}
\toprule
\multirow{2}{*}{Layer} & \multirow{2}{*}{$K$} & \multicolumn{2}{c}{Eff. Labels} & \multirow{2}{*}{Sparsity} & \multirow{2}{*}{\makecell{Recon. \\ Error}} \\
\cmidrule(lr){3-4}
 & & GT & Pred & & \\
\midrule
\rowcolor{sectiongray}
\multicolumn{6}{l}{\textbf{Qwen3}} \\
5  & 128 & 3.47 & 3.33 & 0.703 & 0.126 \\
5  & 256 & 3.13 & 3.07 & 0.771 & 0.098 \\
5  & 512 & 2.64 & 2.67 & 0.830 & 0.080 \\
\cmidrule(lr){1-6}
20 & 128 & 3.74 & 3.12 & 0.708 & 0.126 \\
20 & 256 & 3.13 & 2.75 & 0.774 & 0.099 \\
20 & 512 & 2.78 & 2.52 & 0.830 & 0.081 \\
\cmidrule(lr){1-6}
30 & 128 & 3.48 & 2.39 & 0.782 & 0.054 \\
30 & 256 & 3.21 & 2.15 & 0.835 & 0.041 \\
30 & 512 & 2.83 & 1.92 & 0.877 & 0.032 \\
\midrule
\rowcolor{sectiongray}
\multicolumn{6}{l}{\textbf{LLaVA-1.5}} \\
5  & 128 & 3.65 & 1.80 & 0.643 & 0.155 \\
20 & 128 & 3.66 & 1.76 & 0.722 & 0.103 \\
30 & 128 & 3.63 & 1.74 & 0.739 & 0.109 \\
\bottomrule
\end{tabular}
\caption{Concept decomposition metrics across models, layers, and number of concepts ($K$). Dataset: Wikiart (early modern).}
\label{tab:appendix_concept_decomposition_metrics}
\end{table}
Table~\ref{tab:active_concepts_threshold} shows the effect of the percentile threshold (of non-zero activations) used for binarizing concept activations.
\begin{table}[h!]
\centering
\small
\definecolor{sectiongray}{gray}{0.92}
\setlength{\tabcolsep}{6pt}
\begin{tabular}{ccc}
\toprule
Threshold (\%) & Threshold (Abs.) & \makecell{Avg. Active \\ Concepts} \\
\midrule
\rowcolor{sectiongray}
\multicolumn{3}{l}{\textbf{Qwen3}, Layer $= 30$, $K=128$} \\
0  & 0.00 & 28.65 \\
20 & 0.60 & 22.92 \\
40 & 1.47 & 17.19 \\
60 & 2.92 & 11.46 \\
80 & 6.11 & 5.73  \\
90 & 10.29 & 2.86 \\
95 & 15.48 & 1.43 \\
\bottomrule
\end{tabular}
\caption{Active concepts per image at varying activation thresholds for \texttt{Qwen3}, Layer 30, $K=128$ (the configuration in the main paper). Dataset: Wikiart (early modern).}
\label{tab:active_concepts_threshold}
\end{table}
\subsection{Linear Probing}
\begin{figure}[H]
    \centering
    \includegraphics[width=\columnwidth]{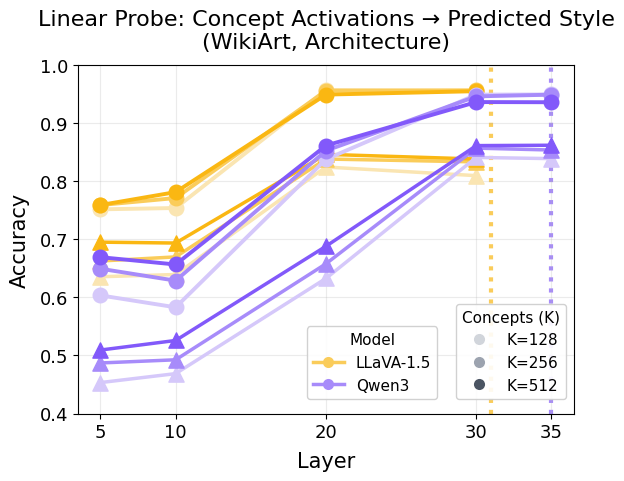}
    \caption{Linear probe accuracy on predicting model output style from concept activations for the Architecture dataset ($\bullet$ raw activations; $\blacktriangle$ binarized activations).
    }
    \label{fig:appendix_linear_probe_architecture}
\end{figure}

\begin{figure}[h!]
    \centering
    \includegraphics[width=\columnwidth]{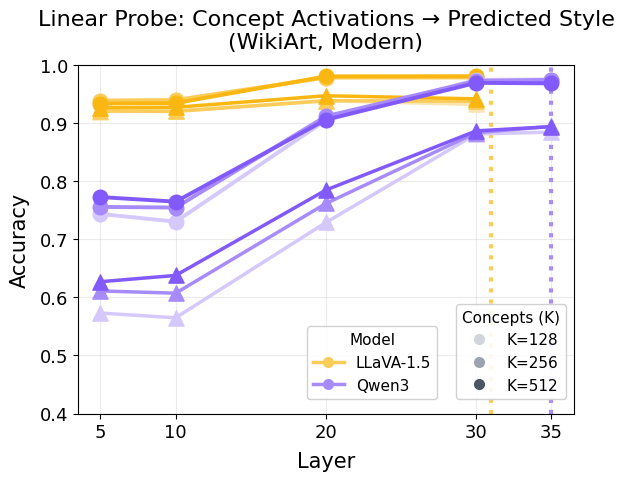}
    \caption{Linear probe accuracy on predicting model output style from concept activations for the Wikiart (modern) dataset.
    }
    \label{fig:appendix_linear_probe_modern}
\end{figure}

Figures~\ref{fig:appendix_linear_probe_architecture} and~\ref{fig:appendix_linear_probe_modern} show linear probing results for our other datasets. \textbf{Note:} these are in the context of patch-level style prediction. For Wikiart (early modern), we additionally experiment with lower numbers of $K=[8, 16, 32, 64]$ (Table \ref{tab:appendix_concept_decomposition_low_k}). We select $K=128$ as it achieves the best metrics while remaining feasible for manual inspection.
\begin{table}[h!]
\centering
\small
\definecolor{sectiongray}{gray}{0.92}
\setlength{\tabcolsep}{6pt}
\begin{tabular}{ccccc}
\toprule
$K$ & Sparsity & \makecell{Recon. \\ Error} & \multicolumn{2}{c}{Linear Probe Acc.} \\
\cmidrule(lr){4-5}
 & & & Raw & Binarized \\
\midrule
\rowcolor{sectiongray}
\multicolumn{5}{l}{\textbf{Qwen3}, Layer $=30$} \\
8  & 0.529 & 0.335 & 0.89 & 0.59 \\
16 & 0.569 & 0.191 & 0.93 & 0.70 \\
32 & 0.642 & 0.115 & 0.94 & 0.74 \\
64 & 0.717 & 0.077 & 0.94 & 0.78 \\
\bottomrule
\end{tabular}
\caption{Concept decomposition metrics for Qwen3 with low numbers of concepts ($K$). Dataset: Wikiart (early modern).}
\label{tab:appendix_concept_decomposition_low_k}
\end{table}
\subsection{Full Image Decomposition}
We select $K=32$, covering the main image motifs in our dataset and 94.7\% of SVD variance.
\section{Concept Labeling Evaluation}
\label{sec:appendix_concept_labeling}
% \kmnote{Not clear what "these" refers to here. I think you mean the labels produced by art historians."}
% AA: changed the language to make clear that our art historian labels are our references
We evaluate our automatically generated labels against human annotator references. For each of the $128$ concepts, we prompt \texttt{GPT-5.4} to produce pairwise judgments of their relative precision, recall, and overall quality when compared to the union of our reference labels.
%We find that 
All three of our methods outperform \textsc{Baseline} (see Table \ref{tab:label_results}), with \textsc{Summary} as the clear winner.
%In considering
%\kmnote{Also not sure what you mean by this last sentence.}
% AA: reworded]
\begin{comment}
\begin{table}[h!]        
  \centering                                        
  \label{tab:label_results}            
  \small
  \begin{tabular}{l ccc ccc ccc}         
  \toprule                        
  & \multicolumn{3}{c}{\textbf{Precision}} & \multicolumn{3}{c}{\textbf{Recall}} & \multicolumn{3}{c}{\textbf{Overall}} \\         
  & \multicolumn{3}{c}{Win Rate} & \multicolumn{3}{c}{Win Rate} & \multicolumn{3}{c}{Win Rate} \\                         
  \cmidrule(lr){2-4} \cmidrule(lr){5-7} \cmidrule(lr){8-10}     
  baseline           & \multicolumn{3}{c}{0.126} & \multicolumn{3}{c}{0.263} & \multicolumn{3}{c}{0.210} \\                     
  similarity          & \multicolumn{3}{c}{0.436} & \multicolumn{3}{c}{0.348} & \multicolumn{3}{c}{0.385} \\                   
  contrastive & \multicolumn{3}{c}{0.547} & \multicolumn{3}{c}{0.434} & \multicolumn{3}{c}{0.486} \\                    
  prompt              & \multicolumn{3}{c}{\textbf{0.891}} & \multicolumn{3}{c}{\textbf{0.956}} & \multicolumn{3}{c}{\textbf{0.919}} \\       
  \bottomrule                          
  \end{tabular}
  \caption{Aggregate win rates from GPT-5.4 pairwise judgments of automatic labels for our $128$ visual concepts, using art historian labels as references.} 
\end{table}
\end{comment}
\begin{table}[h!]        
  \centering          
  \small
  \begin{tabular}{l ccc ccc ccc}         
  \toprule                        
  & \multicolumn{3}{c}{\textbf{Precision}} & \multicolumn{3}{c}{\textbf{Recall}} & \multicolumn{3}{c}{\textbf{Overall}} \\         
  & \multicolumn{3}{c}{Win Rate} & \multicolumn{3}{c}{Win Rate} & \multicolumn{3}{c}{Win Rate} \\                         
  \cmidrule(lr){2-4} \cmidrule(lr){5-7} \cmidrule(lr){8-10}     
  \textsc{Baseline}                    & \multicolumn{3}{c}{0.126} & \multicolumn{3}{c}{0.263} & \multicolumn{3}{c}{0.210} \\                     
  \textsc{Summary}                     & \multicolumn{3}{c}{\textbf{0.891}} & \multicolumn{3}{c}{\textbf{0.956}} & \multicolumn{3}{c}{\textbf{0.919}} \\       
  \textsc{Image-Sim}                   & \multicolumn{3}{c}{0.436} & \multicolumn{3}{c}{0.348} & \multicolumn{3}{c}{0.385} \\                   
  \textsc{Contrastive}                 & \multicolumn{3}{c}{0.547} & \multicolumn{3}{c}{0.434} & \multicolumn{3}{c}{0.486} \\                    
  \bottomrule                          
  \end{tabular}
  \caption{Aggregate win rates from \texttt{GPT-5.4} pairwise judgments of automatic labels for our $128$ visual concepts, using art historian labels as references.} 
  \label{tab:label_results}  
\end{table}

Assessing individual examples, we observe that this performance difference is driven by increased specificity.  In many cases, this is rewarded, as for concept $13$, where the reference label of "hazy soft skies; loose impressionistic brushwork" was best captured by the prompt label of "vague horizon... with indistinct... silhouettes...; low-contrast... blending in muted... tones with soft gradients..."  However, in cases where the concept was deemed less coherent, this specificity leads to errors, as with concept $53$, where our art historians labeled the feature as "no discernible correlation; human figures" while our prompt label was "dimly lit scenes with shadowed... figures in... interiors... often in tense ...moments; moody chiaroscuro ..., strong backlighting..., and loose... texture."  As such, we recommend relying on automatic labels in cases where concepts are coherent. 

\section{Patch-Level Causal Effects}
\label{sec:appendix_causal_analysis}

\begin{figure}[h]
    \centering
    \includegraphics[width=1\columnwidth]{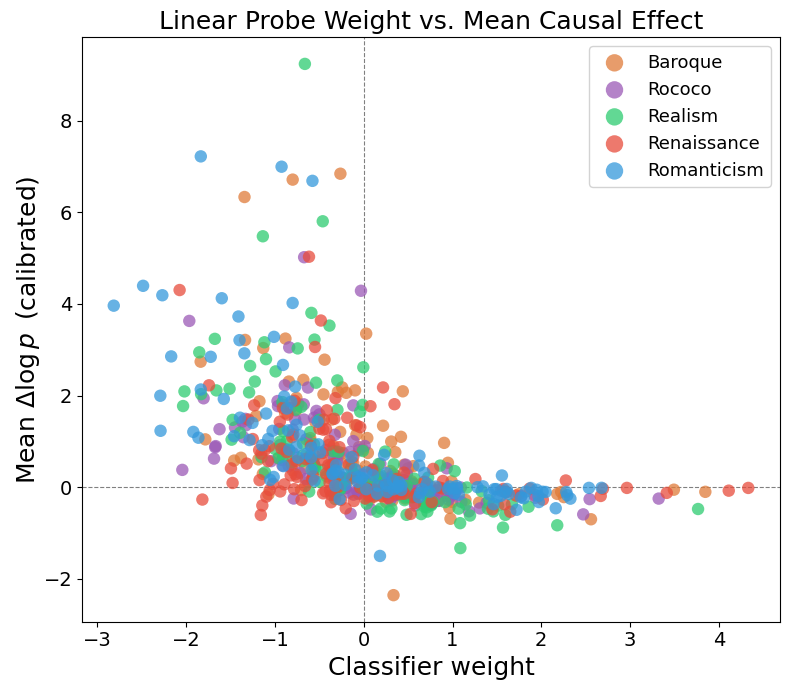}
    \caption{Linear probe weights vs. mean causal effect across styles. Individual points represent a concept, style pair. Spearman's $\rho = -0.714$ ($p = 1.13 \times 10^{-100}$) indicates a strong negative correlation.}
    \label{fig:appendix_causal_effect_baroque}
\end{figure}

\begin{comment}
\begin{equation}
    \log p(s \mid x) = \log p(t_1 \mid x) + \log p(t_2 \mid x, t_1)
    \label{eq:token_probability}
\end{equation}
where $t_{1}, t_{2}$ are the first two tokens of $s$ (e.g., $s=\text{`Realism'}, t_{1}=\text{`Real'}, t_{2}=\text{`ism'}$) and $x$ is the input image and prompt. The two tokens distinguish unique style names. Then, for each of the top three activated concepts, we modify the model's hidden state $\mathbf{h_{L}}$ by subtracting out scaled amounts of the concept direction $\mathbf{v_{i}}$
\end{comment}
 Figure \ref{fig:appendix_causal_effect_baroque} displays the relationship between the linear probe weight of each concept to the target style (x-axis) and the mean causal effect of removing that concept on the prediction of the target style (y-axis). We observe that concepts with higher probe weights causally suppress the model's prediction of the target style when removed.

\end{document}